\documentclass[]{interact}

\usepackage{epstopdf}
\usepackage{subfigure}

\usepackage{makecell}
\usepackage{multirow}
\usepackage{graphicx}
\theoremstyle{plain}

\theoremstyle{definition}

\theoremstyle{remark}

\usepackage{amsmath}
\usepackage[customcolors]{hf-tikz}
\usepackage[hidelinks]{hyperref}
\hypersetup{
	colorlinks   = true, 
	urlcolor     = blue, 
	linkcolor    = blue, 
	citecolor    = red   
}
\usepackage{float}
\usepackage{color,soul}
\usepackage[
style=ieee,
]{biblatex}

\begin{document}

\title{Neural modal ordinary differential equations: Integrating physics-based modeling with neural ordinary differential equations for modeling high-dimensional monitored structures}

\author{
\name{Zhilu Lai\textsuperscript{1,2,*}\thanks{* Correspondence author (formerly at Future Resilient Systems, Singapore-ETH Centre). Email: zhilulai@ust.hk\\
code is available at: \href{https://github.com/zlaidyn/Neural-Modal-ODE-Demo}{https://github.com/zlaidyn/Neural-Modal-ODE-Demo}}, Wei Liu\textsuperscript{3,4}, Xudong Jian\textsuperscript{3,5}, Kiran Bacsa\textsuperscript{3,6}, Limin Sun\textsuperscript{5,7}, and Eleni Chatzi\textsuperscript{3,6}}
\affil{\small
\textsuperscript{1} Internet of Things Thrust, Information Hub, HKUST(GZ), Guangzhou, China.\\
\textsuperscript{2} Department of Civil and Environmental Engineering, HKUST, Hong Kong, China.\\
\textsuperscript{3} Future Resilient Systems, Singapore-ETH Centre, Singapore, Singapore\\
\textsuperscript{4} Department of Industrial Systems Engineering and Management, National University of Singapore, Singapore, Singapore \\
\textsuperscript{5} State Key Laboratory for Disaster Reduction in Civil Engineering, Tongji University, Shanghai, China\\
\textsuperscript{6} Department of Civil, Environmental and Geomatic Engineering, ETH-Z\"urich, Z\"urich, Switzerland\\
\textsuperscript{7} Shanghai Qizhi Institute, Shanghai, China
}}
\maketitle
\begin{abstract}
The order/dimension of models derived on the basis of data is commonly restricted by the number of observations, or in the context of monitored systems, sensing nodes. This is particularly true for structural systems (e.g., civil or mechanical structures), which are typically high-dimensional in nature. In the scope of physics-informed machine learning, this paper proposes a framework -- termed Neural Modal ODEs -- to integrate physics-based modeling with deep learning for modeling the dynamics of monitored and high-dimensional engineered systems. Neural Ordinary Differential Equations -- Neural ODEs are exploited as the deep learning operator. In this initiating exploration, we restrict ourselves to linear or mildly nonlinear systems. We propose an architecture that couples a dynamic version of variational autoencoders with physics-informed Neural ODEs (Pi-Neural ODEs). An encoder, as a part of the autoencoder, learns the abstract mappings from the first few items of observational data to the initial values of the latent variables, which drive the learning of embedded dynamics via physics-informed Neural ODEs, imposing a \textit{modal model} structure on that latent space. The decoder of the proposed model adopts the eigenmodes derived from an eigen-analysis applied to the linearized portion of a physics-based model: a process implicitly carrying the spatial relationship between degrees-of-freedom (DOFs). The framework is validated on a numerical example, and an experimental dataset of a scaled cable-stayed bridge, where the learned hybrid model is shown to outperform a purely physics-based approach to modeling. We further show the functionality of the proposed scheme within the context of virtual sensing, i.e., the recovery of generalized response quantities in unmeasured DOFs from spatially sparse data.
\end{abstract}


\section*{Impact Statement}

We propose Neural Modal ODEs that learn generative dynamical models from spatially sparse sensor data. The proposed method is in the format of dynamical Variational Autoencoders, and we structure the latent space of the measured data using physics-related features (e.g., modal features), allowing physically interpretable architectures. The delivered models are able to reconstruct the full-field structural response, meaning response in unmeasured locations, given limited sensing locations. We believe this proposed method is helpful and meaningful to the community of structural digital twins, model updating, virtual sensing, and structural health monitoring.

\section{Introduction}
Physics-based modeling (or first-principles modeling) forms an essential engineering approach to understand and simulate the behavior of structural systems. Often implemented via the use of finite element methods (FEM) \cite{waisman2010detection,Strømmen2014}, within the context of structural engineering, physics-based modeling is capable of building high-dimensional and high-fidelity models for large and complex civil/mechanical structures. However, such models often suffer from simplified assumptions and approximations, while for the case of monitored operating systems, an established model often fails to reflect a system as is, after possible experience of damaging and deterioration effects. Such limitations can be tackled by means of uncertainty quantification analysis \cite{sankararaman2013bayesian}, or more effectively via feedback from monitoring (sensory) data \cite{farrar2012structural,kamariotis2022value}. The integration of data with physics-based models or physical laws, -- \textit{physics-informed machine learning} \cite{willard2020integrating, karniadakis2021physics,bae2022scientific,zhu2019physics} has grown into an active research area for modeling physical systems in recent years.
 
Beyond their exploitation within a broader science and engineering context \cite{karpatne2017physics,wu2018physics,kashinath2021physics}, physics-informed machine learning has been specifically applied for learning dynamical systems from either simulated or real-world data. This has been pursued in various ways; for instance, by exploiting the automatic differentiation of neural networks to form “custom” activation and loss functions that are tailored to the underlying differential operator \cite{raissi2019physics}, by incorporating Lagrangian dynamics into the Neural Network (NN) architecture \cite{roehrl2020modeling,cranmer2020lagrangian}, by imposing the laws of dynamics as constraints to the network \cite{zhang2020physics2}, or via identification of a sparse set of physics-informative basis functions to establish equations of motion of observed systems \cite{lai2019sparse,lai2020full}. It is further worth noting that a significant tool for fusion lies in the reduction of physics-based models. Notably, Vlachas et al. \cite{Vlachas2022} propose a combination of a long short-term memory network (LSTM) with an autoencoder (AE), jointly referred to as Learning Effective Dynamics, which can be trained on data from simulations of dynamical systems. In a similar context, applied for reduction of nonlinear structural dynamics, Simpson et al \cite{simpson2021machine} combine an LSTM with an AE for delivering fast and accurate simulators of complex high-dimensional structures. In an alternate setting, reduction can efficiently be achieved, while respecting the underlying physics equations, via projection-based methods \cite{qian2020lift,CARLBERG2013623,VLACHAS2021116055}. This yields a powerful framework, which can eventually be combined with data, for instance via use of Bayesian filtering as proposed in \cite{TATSIS2022108558} for the purpose of damage detection and flaw identification. In previous work of part of the authoring team, we delivered hybrid representations that draw from the availability of monitoring data (measurements/observations from the system), which combine a term that reflects our often impartial knowledge of the physics, with a learning term which compensates what our physics representations may not account for, via physics-informed Neural ODEs \cite{lai2021structural} and physics-guided Deep Markov Models (PgDMMs) \cite{liu2021physics}.

Learning a dynamical system essentially boils down to learning a governing function (either in parametric or non-parametric form) that describes the evolution of the ``system's state" over time. We summarize the motivation of this paper as follows. Firstly, in the context of monitoring, the representation of a dynamical system is restricted by the number of sensing nodes. Compared to a model established by physics-based modeling, a data-driven model is often a reduced-order model, typically encompassing contributing modes, which considerably sacrifices the true spatial resolution. Due to this, there often exists an inconsistency between the coordinate spaces of the two models, with the high-dimensional physics-based model (such as a FEM) corresponding to spatially dense DOFs, whilst a data-driven model often reflects a latent space that is expressed in non-physical coordinates \cite{schmid2010dynamic,lusch2018deep,simpson2021machine}. Secondly, the adopted data types are critical to the learning of dynamical systems. If direct measurements of a latent space exist (for example, in representing structural dynamics, displacement and velocity are considered as such latent variables), it is straightforward to learn the dynamics that are inherent to the extracted data. However, this is not the case in practice, as the measured response (data) is most commonly not a direct measurement of the latent variables; for example, when accelerations are available in the context of vibration-based monitoring \cite{ou2021vibration}. With these two aspects in mind, in this paper, we propose a framework that is capable of integrating high-dimensional physics-based models with machine learning schemes for modeling the dynamics of high-dimensional structural systems, with linear or mildly nonlinear behavior. The term ``mildly nonlinear" refers to systems whose response is not significantly different from their linear approximation. Such a discrepancy could be formally quantified using metrics such as the value of the coherence between the input (load) and output (response) signal.


To achieve this, we propose to blend a dynamical version \cite{girin2020dynamical} of a variational autoencoder (VAE) \cite{kingma2013auto}, with a projection basis containing the eigenmodes that are derived from the linearization of a physics-based model, termed as Neural Modal ODEs.
We justify these components in the proposed architecture as follows: (i) the majority of the aforementioned projection-based methods, which commonly rely on proper
orthogonal decomposition (POD \cite{liang2002proper}, have been applied for reduction of nonlinear models/simulators  \cite{Amsallem2015,balajewicz2016projection,abgrall2015robust,vlachas2022coupling,marconia2021enhanced,peherstorfer2016dynamic}).  In this case, we rely on the availability of actual measured data but not simulations of full order models, which may bear with model bias.
To this end, the probabilistic version of autoencoders \cite{hinton1994autoencoders}, i.e., the variational autoencoder (VAE) \cite{kingma2013auto}, is adopted to learn latent representations from data. Our aim is to devise a \textit{generative model}, which is though inferred from data availability, and not a mere observer. In doing so, we exploit data availability in order to infer the initial values of the latent space, in this way boosting the learning of embedded dynamics. This scheme actually falls in the category of non-intrusive model reduction \cite{swischuk2020learning}. In contrast with intrusive model reduction, non-intrusive is data-driven and does not require access to the full order model. (ii) This type of non-intrusive model reduction generally allows for flexibility on the structure of the learned latent space, which need not assume a physically meaningful representation. Since we are interested in monitoring applications, it becomes important to achieve such a physics-based representation, especially for the latent space, since this allows virtual sensing tasks; meaning the inference of structural response in locations that are not directly measured/observed \cite{vettori2022virtual}. To model and structure the dynamics of the reduced-order models (latent dynamics), we herein adopt our previously developed Physics-informed Neural ODEs (Pi-Neural ODEs) \cite{lai2021structural} to impose a modal structure, in which, the dynamics are driven by superposing the modal representations derived from physics-based modeling with a residual term learned by neural networks. This allows accounting for the portion of physics, which remains unaccounted for. (iii) The implemented Pi-Neural ODEs allow for flexibility, as the residual term  adaptively accounts for various discrepancies. In this case, this makes up for the fact that our reduction basis exploits linear eigenmodes. If the system exhibits a mild level of nonlinearity, the resulting discrepancy will be accounted for by the imposed neural network term in the Pi-Neural ODEs.


We validate the efficacy of the proposed Neural Modal ODEs on a numerical example, and an experimental dataset derived from a scaled cable-stayed bridge. Based on the results presented in this paper, the contribution of the study lies in: (i) establishing a generative modeling approach that integrates physics-based modeling with deep learning to model high-dimensional structural dynamical systems, while retaining the format of an ordinary differential equation; (ii) by introducing a physically structured decoder, the model is capable of extrapolating the dynamics to unmeasured DOFs. Such a virtual sensing scheme can be applied to structures where observations are scarce \cite{sun2020structural}; (iii) since this is a generative model, it further has the potential of being implemented within the context of model updating.

    

\section{Neural Modal Ordinary Differential Equations (Nerual Modal ODEs)}

\begin{figure}[H]
    \centering
    \includegraphics[width= .99\linewidth]{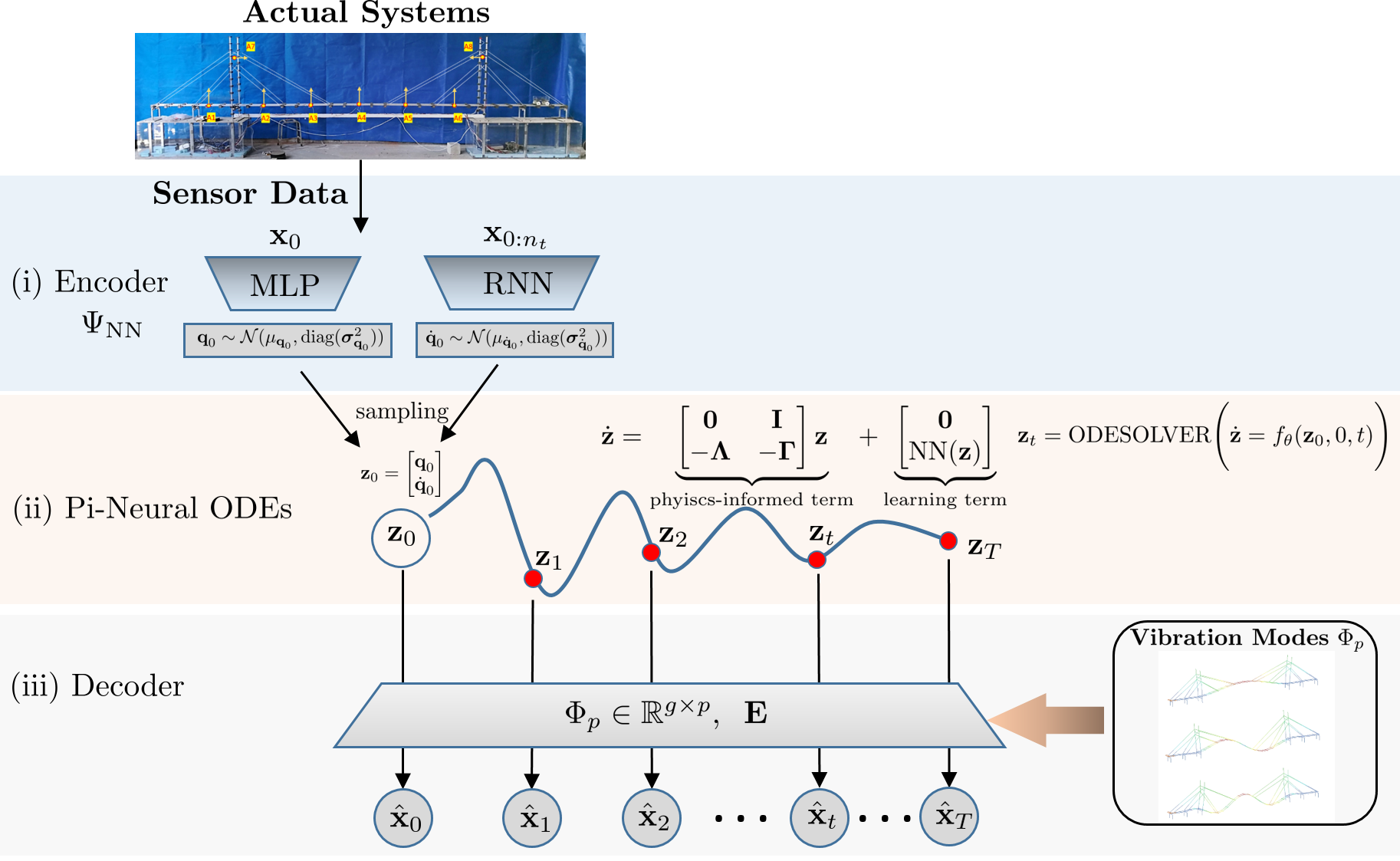}
    \caption{Flow chart of the proposed framework, encompassing a encoder, Pi-Neural ODEs, and a physically structured decoder. The encoder $\Psi_{\text{NN}}$ is comprised of a Multilayer Perceptron (MLP) and a Recurrent Neural network (RNN).}
    \label{fig:flow_chart}
\end{figure}

We summarize the proposed architecture in the flowchart of Figure \ref{fig:flow_chart}, which combines an encoder $\Psi_{\text{NN}}$ and a decoder $\Phi_p$, with  Physics-informed Neural ODEs \cite{lai2021structural} (Pi-Neural ODEs).
The role of the encoder is to perform inference of the initial conditions of the latent variables $\textbf{z}_0$ from a handful of observational data of measured DOFs.

The evolution of the dynamics initiating from $\textbf{z}_0$ is learned and modeled by means of Pi-Neural ODEs. It assumes that a system can be modeled as a superposition of a physics-based modeling term and a learning-based term, where the latter aims to capture the discrepancy between the physics-based model and the actual system. The physics-informed term in this framework adopts a modal representation derived from the eigen-analysis of the structural matrices of the physics-based model. In the case of a nonlinear system, we rely on the linearized portion of the model.

The prediction of latent quantities $\textbf{z}_0, \textbf{z}_1, ... ,\textbf{z}_t, ... ,  \textbf{z}_T$ at time step $t_0, t_1, ...,  t_T$,  obtained from the previous step is mapped back to the full order responses via the decoder, 
and then to the estimated quantities in the original observation space ($\hat{\textbf{x}}_0, \hat{\textbf{x}}_1, ...,\hat{\textbf{x}}_t, ..., \hat{\textbf{x}}_T $) via a selection matrix $\textbf{E}$ (each row is a one-hot row vector), selecting corresponding monitored quantities. This is then compared against the actual measurements to minimize the prediction error, which effectuates the training of the proposed model. The decoder is physically structured, and also derived from the eigen-analysis of the structural matrices. 

In what follows, we offer the details of the formulation of the three outlined components (encoder, Pi-Neural ODEs, and decoder) to the suggested framework.

\subsection{Encoder (Inference Model)}\label{sec:encoder}
Consider an observation (measurement) dataset $\mathcal{D} =\{\textbf{x}^{(i)}\}_{i=1}^{N}$ with $N$ independent sequences of time series data. 
Each sequence reflects a multi-DOF time series record, defined as $\textbf{x}^{(i)} = \{\textbf{x}_0,\textbf{x}_1, ..., \textbf{x}_t, ..., \textbf{x}_T\}^{(i)}$, where the observation vector at time instance $t$, $\textbf{x}_t \in \mathbb{R}^{m}$, reflects $m$ monitored DOFs. When the underlying physics equations are known, the observation $\textbf{x}_t$ at each time instance $t$ can be assumed to be derived from a corresponding latent (state) variable $\textbf{z}_t$, assumed to completely describe the embedded dynamical state. In practice, a common issue is that the latent variables are usually unobserved or only partially observed, via indirect measurements. This limitation is often tackled in prior art via use of an encoder parameterized by a neural network $\Psi_{\text{NN}}$, which is employed to infer the latent variables from observation data.

In delivering such an estimate, we adopt a temporal version \cite{girin2020dynamical} of the variational autoencoder \cite{kingma2013auto}, that has been implemented in existing literature \cite{yildiz2019ode2vae,krishnan2017structured,liu2021physics}. The encoder $\Psi_{\text{NN}}$ can be mathematically described as:
\begin{subequations}
\begin{equation}\label{eq:infer_model_a}
    \Psi_{\text{NN}}(\textbf{z}_0|\textbf{x}_{0:n_t}) =   \Psi_{\text{NN}}\left(\begin{bmatrix}
\textbf{q}_0\\
\dot{\textbf{q}}_0
\end{bmatrix}\Big|\textbf{x}_{0:n_t}\right) 
= \mathcal{N}\left( \begin{bmatrix}
\mu_{\textbf{q}_0}\\
\mu_{\dot{\textbf{q}}_0}
\end{bmatrix},
\begin{bmatrix}
\text{diag}(\bm{\sigma}_{\textbf{q}_0}^2) & \textbf{0}\\
\textbf{0} & \text{diag}(\bm{\sigma}_{\dot{\textbf{q}}_0}^2)\\
\end{bmatrix}
\right),
\end{equation}
where the first few observations from $\textbf{x}_0$ to $\textbf{x}_{n_t}$ (denoted by $\textbf{x}_{0:n_t}$) are used for inferring $\textbf{z}_0$, i.e., $\textbf{z}_0$ is conditioned on $\textbf{x}_0$ to $\textbf{x}_{n_t}$; the latent variables $\textbf{z}_t \in \mathbb{R}^{2p}$ are assumed to have dimension of $2p$, and the output of the encoder is intentionally split into $\textbf{q}_0 \in \mathbb{R}^p$ and $\dot{\textbf{q}}_0 \in \mathbb{R}^p$ that are corresponding to displacement and velocity states, respectively, i.e., $\textbf{z}_0 = \begin{bmatrix}
\textbf{q}_0\\
\dot{\textbf{q}}_0
\end{bmatrix}$.
It is further assumed that the inferred state variable $\textbf{z}_0$ is a stochastic one, which is in this case essential for reflecting uncertainties, and follows a normal distribution, of mean value $\begin{bmatrix}
\mu_{\textbf{q}_0}\\
\mu_{\dot{\textbf{q}}_0}
\end{bmatrix}$ and diagonal covariance matrix $\begin{bmatrix}
\text{diag}(\bm{\sigma}_{\textbf{q}_0}^2) & \textbf{0}\\
\textbf{0} & \text{diag}(\bm{\sigma}_{\dot{\textbf{q}}_0}^2)\\
\end{bmatrix}$. It should though be noted that it is common to model uncertainty in structural systems, which are subjected to random environmental influences, using a normal distribution. For most of dynamical variational autoencoders frameworks, which are adopted in the context of modeling dynamical systems with uncertainty, the inherent uncertainties are accounted for via use of normal distributions, as summarized in the work of Girin et al. \cite{girin2020dynamical}.

In practice, $\Psi_{\text{NN}}$ is comprised of a feed-forward neural network (Multilayer Perceptron, MLP) and a Recurrent Neural network (RNN). We assume that the displacement quantity $\textbf{q}_0$ only depends on $\textbf{x}_0$, per the assumption adopted in \cite{yildiz2019ode2vae}: 
\begin{equation}
    (\mu_{\textbf{q}_0}, \bm{\sigma}_{\textbf{q}_0}^2) = \text{MLP}(\textbf{x}_0),
\end{equation}
The output of this MLP is a stochastic variable of mean $\mu_{\textbf{q}_0}$ and variance $\bm{\sigma}_{\textbf{q}_0}^2$; the velocity quantity $\dot{\textbf{q}}_0$ is inferred from the first leading observations $\textbf{x}_{0:n_t}$, thus a RNN is implemented to take $\textbf{x}_0, \textbf{x}_1, ..., \textbf{x}_{n_t}$ into account:
\begin{equation}
  (\mu_{\dot{\textbf{q}}_0}, \bm{\sigma}_{\dot{\textbf{q}}_0}^2) = \text{RNN}(\textbf{x}_{0:n_t}),
\end{equation}
where the output of the RNN is a stochastic variable of mean $\mu_{\dot{\textbf{q}}_0}$ and variance $\bm{\sigma}_{\dot{\textbf{q}}_0}^2$; $n_t$ need not necessarily reflect a large number, larger $n_t$ might dilute the inference of the velocity quantity; for instance, based on empirical trial, in our implementations $n_t = 10$. 
\end{subequations}
Once the normal distribution defined in Eq.\eqref{eq:infer_model_a} is derived, one can sample $\textbf{z}_0$ from this distribution, and use it for computing the evolution of the latent dynamics over time. We use $\bm{\theta}_{\text{enc}}$ to denote all the parameters used in the $\Psi_{\text{NN}}$, i.e., all the hyper-parameters involved in the formulation of the MLP and RNN architectures.

\subsection{Modeling Latent Dynamics via Physics-informed Neural ODEs}\label{sec:model_dyn}
There are generally two strategies in terms of how the temporal dependence between states $\textbf{z}$ can be modeled. The first strategy is to use a discrete-time model to describe the embedded dynamics, where the first-order Markovian property is assumed. A popular example, in the Deep Learning context, can be found in the deep Markov models \cite{krishnan2015deep, krishnan2017structured,liu2021physics}. An alternative lies in adopting continuous models, usually in the form of differential equations, to describe the temporal dependence embedded in the data. The neural ordinary differential equations (Neural ODEs) \cite{chen2018neural} form a recently proposed tool that parameterizes the governing differential equations by feed-forward neural networks in a continuous format. A specific merit of a continuous modeling approach is that non-equidistant sequential data can be used for training the model. As the Neural ODEs effectively represent a differential equation construct, the trained model can, in turn, be used as a generative model, meaning as a model which can predict the system response given initial conditions or external excitation.

In previous work of the authors \cite{lai2021structural}, we introduced a physics-informed Neural ODEs (Pi-Neural ODEs) scheme, assuming that a system can be modeled as a superposition of a physics-based modeling term and a learning-based term, where the latter aims to capture the discrepancy between the physics-based model and the actual system. A similar scheme is further discussed in \cite{wagg2020digital} for application within the context of digital twinning, as the learning-based term allows for adaptation. The scheme is formally described as follows:
\begin{equation}
   \dot{\textbf{z}} = f_{\bm{\theta}_{\text{dyn}}}(\textbf{z}) = f_{\text{phy}}(\textbf{z}) +  f_{\text{NN}}(\textbf{z}),
\end{equation}
where $f_{\text{phy}}(\textbf{z})$ is a physics-based model, which can be built by leveraging the best possible knowledge of the system; $f_{\text{NN}}(\textbf{z})$ is the learning-based model that is materialized as a neural network function of $\textbf{z}$. It is noted that the former term $f_{\text{phy}}(\textbf{z})$ is of a fixed and pre-assigned structure, while the latter term is adjustable during the process of training the model. The parameter vector $\bm{\theta}_{\text{dyn}}$, reflects the set of hyper-parameters involved in the neural network representation $f_{\text{NN}}(\textbf{z})$.

In this paper, we adopt this modeling scheme for use within a reduced order modeling (ROM) setting, to model the latent dynamics of a high-dimensional system. We restrict ourselves in this initiating effort to the modeling of linear or mildly nonlinear systems. The mildly nonlinear system we refer to in this paper is that the system can be well approximated by the linearization of the system -- the first-order Taylor expansion.

In such a case, an approximation of the dynamics can be derived through the solution of an eigenvalue problem of the structural matrices of the physics-based model (in the case of a nonlinear system, we rely on the linearized part), and is reflected in the following decoupled low-dimensional linearized form:
\begin{subequations}
\begin{equation}\label{eq:latent_dynamics}
  \begin{bmatrix}
  \dot{\textbf{q}}\\
  \ddot{\textbf{q}}
  \end{bmatrix}   = 
  \begin{bmatrix}
  \textbf{0} & \textbf{I}\\
  -\bm{\Lambda}  &  -\bm{\Gamma}
  \end{bmatrix}
  \begin{bmatrix}
\textbf{q}\\
  \dot{\textbf{q}}
  \end{bmatrix},
\end{equation}
where, 
\begin{equation}
    \bm{\Lambda} = \begin{bmatrix}
    \omega_1^2 & & \\
    & \omega_2^2 & &  \\
    &  & \ddots &  \\
     &  &  & \omega_p^2  \\
    \end{bmatrix} 
    \;\;
   \bm{\Gamma} =    \begin{bmatrix}
    2\xi_1\omega_1 & & \\
    & 2\xi_2\omega_2  & &  \\
    &  & \ddots &  \\
     &  &  & 2\xi_p\omega_p   \\
    \end{bmatrix},
\end{equation}
where $\bm{\Lambda}$ and $\bm{\Gamma}$ are both diagonal matrices;  $\omega_1, \omega_2, ..., \omega_p$ are the first $p$ leading natural frequencies (the first $p$ maximum frequencies in a descending order) that are retrieved from an eigen-analysis of an a priori available physics-based model; $\xi_1, \xi_2, ..., \xi_p$ are the corresponding modal damping ratios; $\textbf{I} \in \mathbb{R}^{p\times p}$ denotes the identity matrix.  

Our premise is that the physics-based model in Eq.\eqref{eq:latent_dynamics} does not fully represent the actual system, which implies that the model-derived modal parameters can be different from the parameters that describe the actual operating system as-is, or that additionally, further to the parameters, the structure of the model is lacking. The latter implies that certain mechanisms are not fully understood and are, thus, modeled inaccurately, for instance, mechanisms related to nonlinearities or damping. To account for such sources of error or discrepancies, we add a learning-based term to model the dynamics that are unaccounted for, with Eq.\eqref{eq:latent_dynamics} now defined as:
\end{subequations}
\begin{equation}\label{eq:Pi-Neural}
   \dot{\textbf{z}}  = 
  \begin{bmatrix}
  \textbf{0} & \textbf{I}\\
  -\bm{\Lambda} &  -\bm{\Gamma}
  \end{bmatrix}
\textbf{z}  + 
  \begin{bmatrix}
  \textbf{0}\\
  \text{NN}\left(
\textbf{z} 
  \right)
  \end{bmatrix} \;\; \text{with} \;\; \textbf{z}(0) = \textbf{z}_0,
\end{equation}
where $\textbf{z} =  \begin{bmatrix}
  \textbf{q}\\
  \dot{\textbf{q}}
  \end{bmatrix}$; \text{NN} represents a feed-forward neural network that is a function of $\textbf{z}$. It is noted that the structure presented in Eq.\eqref{eq:Pi-Neural} has the potential of breaking the fully decoupled structure, which is defined by the first term. This is in fact welcomed since the hypothesis of fully decoupled damping matrices, relating to a Rayleigh viscous damping assumption \cite{craig2006fundamentals}, is a known source of modeling discrepancies for real-world systems \cite{satake2003damping}. The learning-based term $\text{NN}(\textbf{z})$ is thus added to account for possible sources of inconsistency and error. In this physics-informed architecture, during training, the estimated gradients are obtained as the sum of the corresponding gradients derived from the physics-based and learning-based terms. Since the gradients from the physics-based term are fixed, only the gradients of the learning-based term are to be estimated. The combined gradients are restricted in a regime that is closer to the true function's gradients. Appendix \ref{appendix_A} further elaborates on the benefit of this physics-informed architecture, which boosts the search for the governing equations close to the actual systems.
 
The Physics-informed Neural ODE Eq.\eqref{eq:Pi-Neural} governs the evolution of the dynamics. The dynamics of $\textbf{z}(t)$ can be solved by numerically integrating $\textbf{z}(t) = \int_{t_0}^{t} f_{\bm{\theta}_{\text{dyn}}}(\textbf{z})dt$ from $t_0$ to $t$ given initial conditions $\textbf{z}_0$, with the estimate of the latent state vector $\textbf{z}(t)$ at each time $t$ offered as:
\begin{equation}\label{eq:transition_}
    \textbf{z}(t) = \text{ODESOLVE}(f_{\bm{\theta}_{\text{dyn}}},\textbf{z}_0,t_0,t),
\end{equation}
where ODESOLVE reflects the chosen numerical integration scheme, with Runge-Kutta methods comprising a typical example of such solvers. The dynamics of the latent state $\textbf{z}$, with realization of $\textbf{z}_0, \textbf{z}_1, ..., \textbf{z}_t,..., \textbf{z}_T$ (where $\textbf{z}_t = \begin{bmatrix}
\textbf{q}_t\\
\dot{\textbf{q}}_t
\end{bmatrix}$), are thus computed at each time step, and can be subsequently fed into the decoder model to reconstruct the full field response, as described in what follows. 

\subsection{Decoder}\label{sec:decoder}
In the case of a linear dynamical system, the full-order response $\textbf{x}^{\text{full}}_t \in \mathbb{R}^g$ comprises a modal representation of $\textbf{x}^{\text{full}}_t \approx \Phi_p\textbf{q}_t \;\; (\Phi_p \in \mathbb{R}^{g\times p}; \textbf{q}_t \in \mathbb{R}^{p}; p \leq g)$, where $\Phi_p$ is the truncated eigenvector matrix, i.e., the leading $p$ columns of full-order eigenvector matrix $\Phi$ (corresponding to the largest $p$ eigenvalues). 

As illustrated in Figure \ref{fig:flow_chart}, an estimate of the evolution of the latent state over time $\textbf{z}_0, \textbf{z}_1, ... , \textbf{z}_T$ can be obtained by solving the Pi-Neural ODEs via Eq.\eqref{eq:transition_}. It is noted that, within the structural dynamics context, important measurable quantities such as accelerations $\ddot{\textbf{q}}$ can further be computed on the basis of the governing Eq.\eqref{eq:Pi-Neural}: $\ddot{\textbf{q}} = \begin{bmatrix}
-\bm{\Lambda} & -\bm{\Gamma}
\end{bmatrix}\textbf{z} + \text{NN}(\textbf{z})$. 
Thus, beyond the latent states $\textbf{q}$, $\dot{\textbf{q}}$, we can derive further response quantities of interest, such as the acceleration  $\ddot{\textbf{q}}$. 

Each response quantity can be respectively emitted to the corresponding full-order response vector (involving all structural DOFs) via the decoder $\Phi_p$ ($\mathbb{R}^p \rightarrow \mathbb{R}^g$):
\begin{subequations}
\begin{equation}\label{eq:emission}
\begin{split}
   & \text{displacement:}  \;\;\; \textbf{x}^{\text{full}}_t = \Phi_p(\textbf{q}_t),   \\
   & \text{velocity:} \;\;\; \dot{\textbf{x}}^{\text{full}}_t = \Phi_p(\dot{\textbf{q}}_t),  \\
      & \text{acceleration:} \;\;\; \ddot{\textbf{x}}^{\text{full}}_t = \Phi_p(\ddot{\textbf{q}}_t),  \;\;\; (t = 0,1,..., T) \\
\end{split}
\end{equation}
where $\textbf{x}^{\text{full}}_t$, $\dot{\textbf{x}}^{\text{full}}_t$, and $\ddot{\textbf{x}}^{\text{full}}_t$ denote the reconstructed full-order displacement, velocity, and acceleration, respectively. It is noted that further response quantities of interest, such as potentially strains, can be inferred due to availability of a FEM model.

We can only measure a limited of DOFs, $\textbf{x}_t \in \mathbb{R}^m$ (we use $\textbf{x}_t$ to denote measured quantities while $\hat{\textbf{x}}_t$ denoting the corresponding estimated quantities), via use of appropriate sensors, which form a subset of the full response vector:
\begin{equation}\label{eq:emission2}
    \hat{\textbf{x}}_t = \textbf{E}\begin{bmatrix}
    \textbf{x}^{\text{full}}_t\\
    \dot{\textbf{x}}^{\text{full}}_t\\
    \ddot{\textbf{x}}^{\text{full}}_t
    \end{bmatrix},
\end{equation}
\end{subequations}
where $\textbf{E} \in \mathbb{R}^{m \times 3g}$ is a selection matrix (each row is a one-hot row vector), selecting corresponding monitored quantities; $\hat{\textbf{x}}_t$ can represent an extended set of the estimated observations, which can correspond to displacement, velocity, acceleration, or further computable response quantities (such as strains). Since we only consider mild nonlinearity, we rely on the observability of the linearized part of the system, where classical observability theory \cite{kalman1960general} can be applied to analyze the observability -- estimating the full state vector from limited measurements.

The architecture of the proposed framework essentially comprises a sequential version of the Variational Autoencoder (VAE), exploiting the presence of an underlying low-dimensional latent representation in the observed dynamics. In the original VAE, the decoder is parameterized by a neural network without regularization, which flexibly fits the training data, without necessarily embodying a physical connotation. From an engineering perspective, however, it would be beneficial if the decoder is bestowed with a direct linkage to physical DOFs. One way to achieve this is to seed the modal shape information, computed from physics-based models, which carries within it the spatial information of how each element/node in $\textbf{x}$ is interconnected. Therefore, we forcibly implement eigenmodes $\Phi_p$ as the decoder for emitting the latent variables to the observation space. $\Phi_p = [\phi_1, \phi_2, ...., \phi_p]$, where each column represents a single eigenmode, can be derived from the structural matrices of the physics-based full order model, e.g. a FE model. It is noted that $\Phi_p$ is assumed to be time-invariant, thus reflecting an invariant encoding of the spatial relationship between structural DOFs. However, the residual term $\text{NN}(\textbf{z})$ in the Pi-Neural ODE in Eq.\eqref{eq:Pi-Neural} adaptively accounts for discrepancies that stem from mild nonlinearities, which would also violate the assumption of invariance. We remind that, in section \ref{sec:model_dyn}, a decoupled structure is adopted as a prior model to encourage the model to mimic the process of a modal-decomposition-reconstruction.

It is worth mentioning that, in this framework, the encoder process can be viewed as the transformation from full order physical coordinates to modal coordinates $\Psi_{\text{NN}}: \textbf{x} \rightarrow \textbf{z}$. In real scenarios that involve weakly nonlinear systems, this can be thought of as a ``modal-like" coordinate as the learning term $\text{NN}(\textbf{z})$ can violate the decoupled structure, while the decoder is viewed as the operator which enables the transformation from the modal coordinates' space to the measured physical coordinates ($\Phi_p: \textbf{z} \rightarrow \textbf{x}$).


\subsection{Loss Function}
For the purpose of training the suggested Neural Modal ODE models, which capitalize on the availability of physics information and data, we calculate the measurement prediction error. The model delivers an estimate  $\hat{\textbf{x}}_{0:T}$ of the measured response quantities $\textbf{x}_{0:T}$, which in turn allows to minimize the error between the predicted and actual observations, to train the model. The training of the encoder, decoder, and latent dynamic models are performed simultaneously, and the loss function of the framework is given as:
\begin{equation}
    \mathcal{L}(\bm{\theta}; \textbf{x}) = \mathcal{L}\Big\{  \text{DECODER}\big[\text{ODESOLVE}(f_{\bm{\theta}_{\text{dyn}}}, \Psi_{\text{NN}}(\textbf{x}_{0:n_t}),t_0,T) \big]\Big\},
\end{equation}
where $\bm{\theta} = \bm{\theta}_{\text{enc}}\cup \bm{\theta}_{\text{dyn}}$ are all the parameters involved in the deep learning model; $\textbf{x}_{0:n_t}$ is the first $\textbf{x}_0$ to $\textbf{x}_{n_t}$ data fed into the encoder $\Psi_{\text{NN}}$; $\textbf{x}_{0:T}$ is the whole sequence of the data set used for the decoder; the notation DECODER denotes the process given in Eq.\eqref{eq:emission} and \eqref{eq:emission2}. 

In the VAE formulation \cite{kingma2013auto}, the loss function $\mathcal{L}$ is used to \textit{maximize} a variational lower bound of the data log-likelihood $\log p(\mathbf{x})$; here \textbf{x} is short for $\textbf{x}_{0:T}$. Using the variational principle with the inference model $\Psi_{\text{NN}}(\mathbf{z}_0|\mathbf{x}_{0:n_t})$, which is only used to infer the initial condition of $\textbf{z}_0$, the evidence lower bound (ELBO) of the data log-likelihood, which is the loss function, is given as follows:
\begin{equation}\label{ELBO_fac}
\mathcal{L}(\bm{\theta};\mathbf{x})=\sum_{t=0}^T\Big\{\mathbb{E}_{\Psi_{\text{NN}}(\mathbf{z}_0|\mathbf{x}_{0:n_t})}[\log p(\mathbf{x}_t|\mathbf{z}_t)]
-\mathbb{E}_{\Psi_{\text{NN}}(\mathbf{z}_0|\mathbf{x}_{0:n_t})}\big[\text{KL}\big(\Psi_{\text{NN}}(\mathbf{z}_0|\mathbf{x}_{0:n_t})||p(\mathbf{z}_0)\big)\big]\Big\},
\end{equation}
where KL stands for the Kullback–Leibler divergence; a statistical measure that evaluates the closeness of two probability distributions $p_1$ and $p_2$, defined as $\text{KL}(p_1(\mathbf{z})||p_2(\mathbf{z})) := \int p_1(\mathbf{z}) \log\frac{p_1(\mathbf{z})}{p_2(\mathbf{z})} d\mathbf{z}$. In the loss function, the first term $\sum_{t=0}^T\mathbb{E}_{\Psi_{\text{NN}}(\mathbf{z}_0|\mathbf{x}_{0:n_t})}[\log p(\mathbf{x}_t|\mathbf{z}_t)]$ evaluates the reconstruction accuracy: $\textbf{z}_0$ is sampled from the distribution given in Eq.\eqref{eq:infer_model_a}, and with this given initial condition, one can compute the predicted $\hat{\textbf{x}}_t  \sim \mathcal{N}(\hat{\bm{\mu}}_t, \hat{\bm{\Sigma}}_t) \;
\; (t = 0,1, ... ,T)$ via the latent dynamics model in Eq.\eqref{eq:transition_} followed by the decoder. Thus, this term can be computed as $\sum_{t=0}^T\log p(\mathbf{x}_t)$ given $\textbf{z}_0 \sim \Psi_{\text{NN}}(\mathbf{z}_0|\mathbf{x}_{0:n_t})$, and $\log p(
\textbf{x}_t)$ has an analytical form when $p(\textbf{x}_t)$ follows a normal distribution: 
\begin{equation}\label{eq:likelihood}
\log p(\mathbf{x}_t)=-\frac{1}{2}\Big[\log|\hat{\bm{\Sigma}}_t|+(\textbf{x}_t-\hat{\bm{\mu}}_t)^T\hat{\bm{\Sigma}}_t^{-1}(\textbf{x}_t-\hat{\bm{\mu}}_t) +d_\textbf{x}\log(2\pi)\Big],
\end{equation}
which is the log-likelihood, and the training of the model is expected to maximize this likelihood given the actual observation data $\textbf{x}_t$; $d_\textbf{x}$ is the dimension of $\textbf{x}_t$. 

The second term $-\sum_{t=0}^T
\mathbb{E}_{\Psi_{\text{NN}}(\mathbf{z}_0|\mathbf{x}_{0:n_t})}\big[\text{KL}\big(\Psi_{\text{NN}}(\mathbf{z}_0|\mathbf{x}_{0:n_t})||p(\mathbf{z}_0)\big)\big]$ evaluates the closeness of the inferred initial condition with a prior distribution $p(\textbf{z}_0)$. In practice, $p(\mathbf{z}_0)$ can be assumed as a normal distribution $\mathcal{N}(\textbf{0},\textbf{I})$ if no further prior knowledge is given. The KL terms acts as a penalty term when the inferred initial value is distant from the prior distribution. This term can be alternatively computed as $-\sum_{t=0}^T \text{KL}\big(\Psi_{\text{NN}}(\mathbf{z}_0|\mathbf{x}_{0:n_t})||p(\mathbf{z}_0)\big)$ given $\textbf{z}_0 \sim \Psi_{\text{NN}}(\mathbf{z}_0|\mathbf{x}_{0:n_t})$. $\text{KL}(p_1(\mathbf{z})||p_2(\mathbf{z}))$ is described by an analytical formula when both $p_1(\textbf{z})$ and $p_2(\textbf{z})$ are normal distributions and $p_2(\textbf{z}) \sim \mathcal{N}(\textbf{0},\textbf{I})$:
\begin{equation}
    \text{KL}\Big(\Psi_{\text{NN}}(\mathbf{z}_0|\mathbf{x}_{0:n_t})||p(\mathbf{z}_0)\Big) = -\log|\text{diag}(\bm{\sigma}_{\textbf{z}_0})| + \frac{||\bm{\sigma}_{\textbf{z}_0}||^2 + ||\bm{\mu}_{\textbf{z}_0}||^2}{2} - \frac{d_\textbf{z}}{2},
\end{equation}
in which, $\bm{\sigma_{\textbf{z}_0}} = \begin{bmatrix}
\bm{\sigma_{\textbf{q}_0}}\\
\bm{\sigma_{\dot{\textbf{q}}_0}}
\end{bmatrix}$; $\bm{\mu_{\textbf{z}_0}} = \begin{bmatrix}
\bm{\mu_{\textbf{q}_0}}\\
\bm{\mu_{\dot{\textbf{q}}_0}}
\end{bmatrix}$; $|\cdot|$ is the determinant of a matrix; $||\cdot||$ is the modulus of a vector; $d_{\textbf{z}}$ is the dimension of $\textbf{z}$.  
    

\subsection{Prediction of learned dynamics}\label{sec:pred}
The completion of the training process results in the definition of the hyper-parameter sets $\bm{\theta}_{\text{enc}}$ and $\bm{\theta}_{\text{dyn}}$. This delivers an encoder $\Psi_{\text{NN}}$  together with a learned dynamic model $\dot{\textbf{z}} = f_{\bm{\theta}_\text{dyn}}(\textbf{z})$, which retains the structure of differential equations. Eqs.\eqref{eq:transition_} and \eqref{eq:emission} can be used for predicting the dynamics given an initial state $\textbf{z}_0$. $\textbf{z}_0$ can be either be inferred from the observation dataset via the learned encoder $\Psi_{\text{NN}}$, or -- when using the derived model as a generative model -- the modeler can assign other specific values for the initial condition $\textbf{z}_0$. 

For those readers that are interested in reusing the developed algorithms, a demonstrative implementation in Python, reproducing all steps from Sections \ref{sec:encoder} to \ref{sec:pred}, will be made available at: \href{https://github.com/zlaidyn/Neural-Modal-ODE-Demo}{https://github.com/zlaidyn/Neural-Modal-ODE-Demo}, including both linear and nonlinear cases of a demonstrative example introduced in the next section.

\section{Demonstrative Example of a 4-DOF Structural System}\label{sec:num_exp}
In this section, we implement the proposed framework on a simulated 4-DOF structural systems. The structural system is governed by the following differential equations:
\begin{subequations}
\begin{equation}\label{eq:4dof}
    \textbf{M}\ddot{\textbf{x}} + \textbf{C}\dot{\textbf{x}} + \textbf{K}\textbf{x} + \begin{bmatrix}
    0\\
    0\\
    0\\
    k_n x_1^3
    \end{bmatrix} = \textbf{0},
\end{equation}
where the displacement vector $\textbf{x} = [x_1, x_2, x_3, x_4]^T$; the mass matrix $\textbf{M} = \text{diag}(m_1,m_2,m_3,m_4)$, and $m_1 = 1, m_2 = 2, m_3 = 3, m_4 = 4 $; the damping matrix $\textbf{C} = \text{diag}(c_1,c_2,c_3,c_4)$, and $c_1 = c_2 = c_3 = c_4 = 0.1 $; and the stiffness matrix
\begin{equation}
\textbf{K} = 
\begin{bmatrix}
&k_1+k_2 &-k_2 &0 &0\\
&-k_2 &k_2+k_3 &-k_3 &0\\
&0 &-k_3 &k_3+k_4 &-k_4\\
&0 &0 &-k_4 &k_4
\end{bmatrix},
\end{equation}
\end{subequations}
where $k_1 = 1, k_2 = 2, k_3 = 3, k_4 = 4$. To fully demonstrate the capability of the proposed framework for both linear and nonlinear structural systems, we test three different cases with increasing nonlinearity $k_n=0.0 \; (\text{linear case}),0.5$, and 1.0, respectively. The linear portion of the three cases are set to be the same and the only variation lies in the coefficient $k_n$ as the nonlinear term.

A total number of 1,000 realizations with randomized initial conditions from a standard normal distribution are generated for each case (the randomization is identical for each case of $k_n$). As mentioned, we assume that in the here presented application scenarios only a limited subset of the full-order system response quantities are available. In this example, only the displacement of the fourth DOF ($x_4$) and the accelerations of the first, third and fourth DOFs ($\ddot{x}_1, \ddot{x}_3, \ddot{x}_4$) are measured. While it is feasible to implement the framework with acceleration measurements only, the accounted displacement of a single DOF is here used to alleviate possible drifting effects that occur in the reconstructed full state.
The first $n_0$ to $n_t=10$ samples of the sequence are used for the RNN in the encoder to infer the initial latent velocity. As for the decoder $\Phi_p$, we make use of the first $p=4$ modes obtained via an eigen-analysis of the structural matrices of the physics-based model, thus forming an 8-dimensional latent state. The implementation details are listed in Table \ref{tab:NNs_num}. The models are trained on the dataset of the first 800 realizations and tested on the remaining 200 realizations.

\begin{table}[H]
 \small
    \centering
    \caption{Implementation details for the numerical study}
    \begin{tabular}{c c c c c}
   
    \hline\hline
      & \multicolumn{2}{c}{Encoder} & Modeling latent dynamics & Decoder \\
        \hline
    &  \makecell{RNN($\textbf{x}_{0:n_t}$)\\\small{$n_t = 10$}} &  MLP($\textbf{x}_0$)   &    $\text{NN}(\textbf{z})$ in Eq.\eqref{eq:Pi-Neural} & $\Phi_p$ \\
  \hline
\makecell{no. of \\hidden layers} & 1 & 2 & 2 &\multirow{2}{*}{\makecell{invariant,\\\small{$\Phi_p = [\phi_1, \phi_2, \phi_3, \phi_4]$}}   }\\  
\makecell{no. of neurons in \\each hidden layer}  & 32 & 128 & 128 & \\  
  \hline
    \end{tabular}
    \label{tab:NNs_num}
\end{table}

\begin{figure}[H]
    \centering
    \includegraphics[width = 0.5\textwidth]{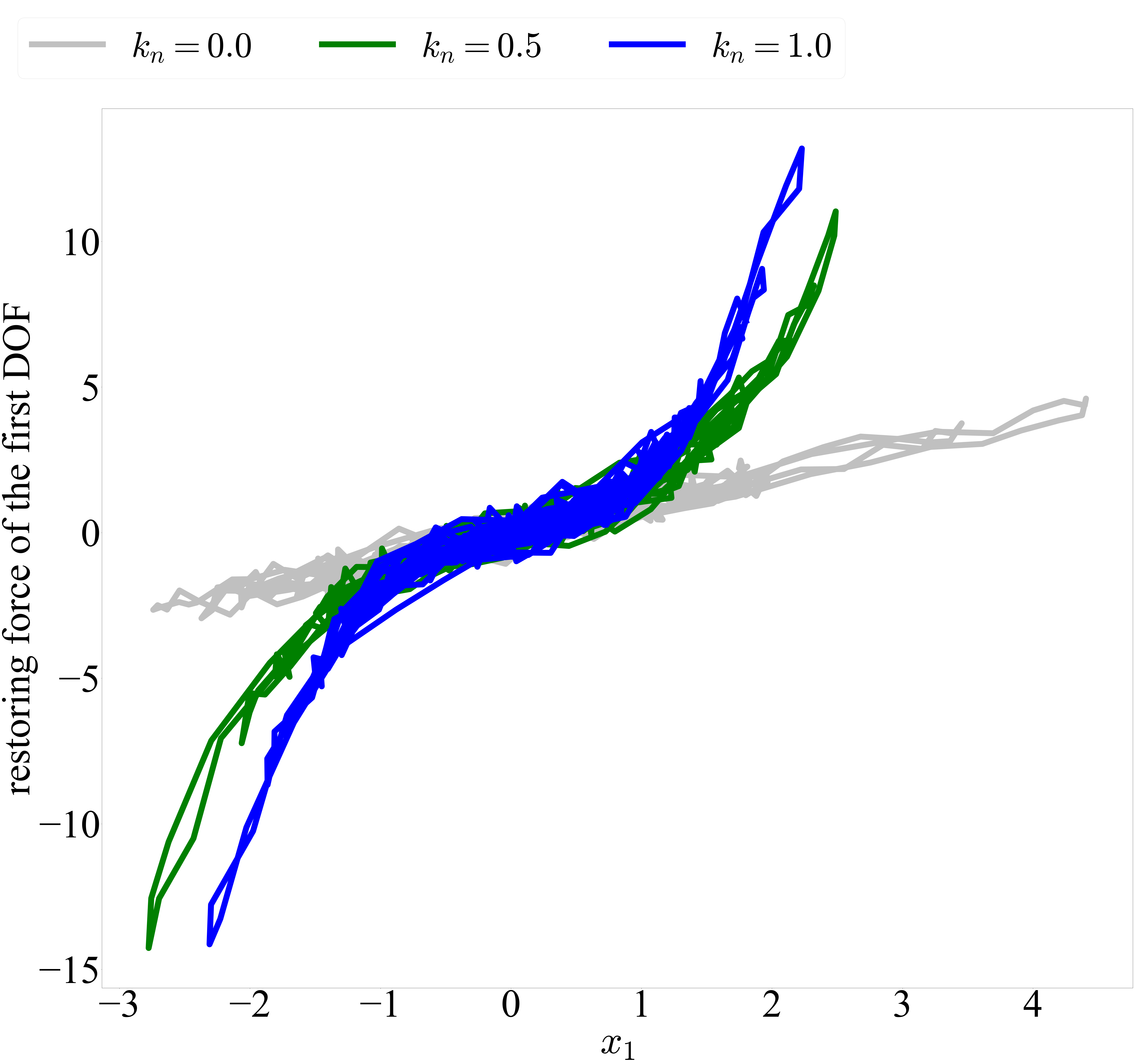}
    \caption{The force-displacement loops of the 1st DOF of the reference system for different values of the nonlinear coefficient $k_n$.}
    \label{fig:3non}
\end{figure}

Figure \ref{fig:3non} shows the  force-displacement loops of the 1st DOF of the reference system for different values of the nonlinear coefficient $k_n$. It indeed reveals that the simulated data delivers different levels of nonlinearity and the measured data are contaminated with noise.

\begin{figure}[H]
    \subfigure[$k_n = 0$]{\label{fig:kn_0}\includegraphics[width=0.95\linewidth]{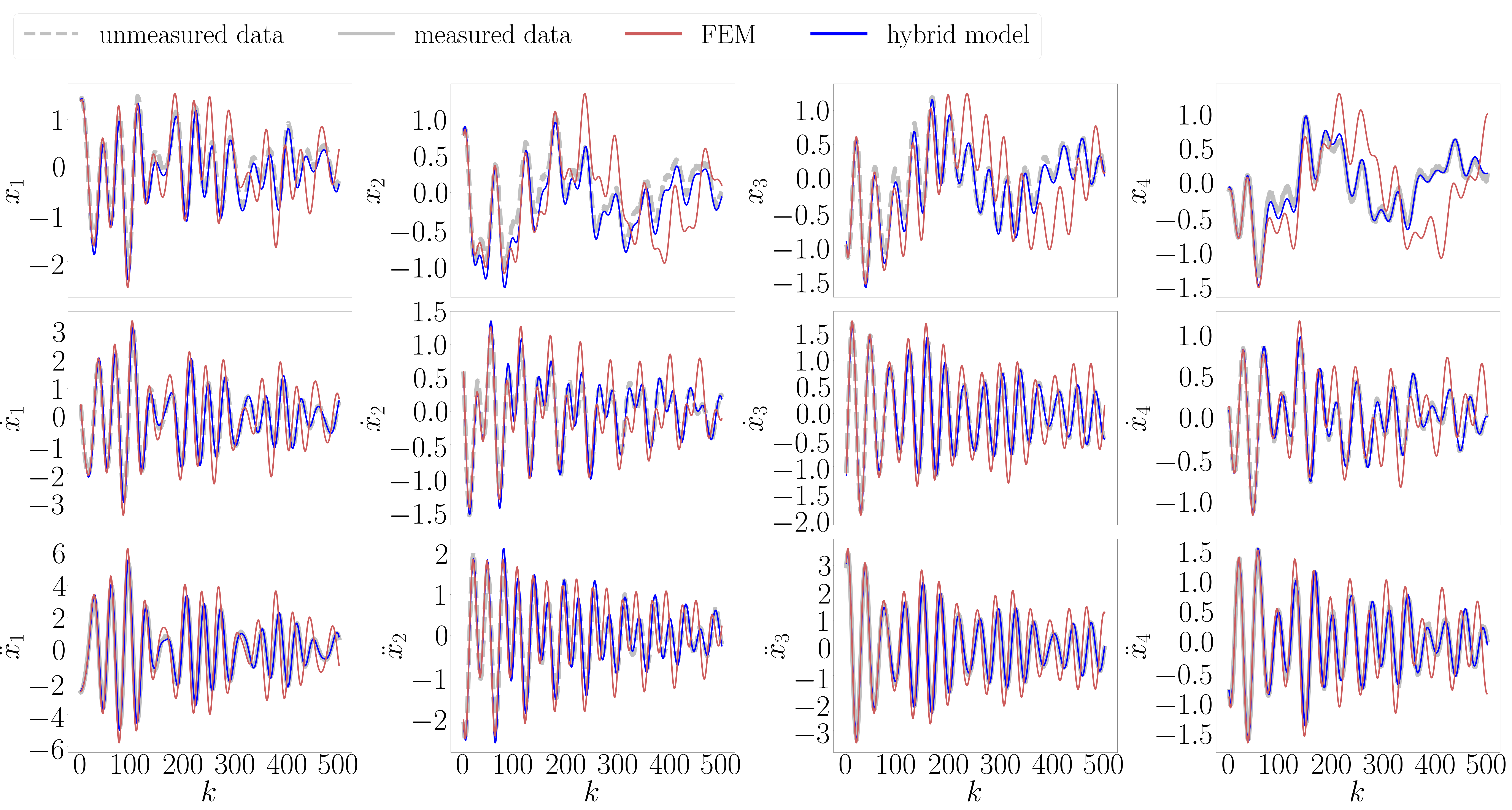}}
    \subfigure[$k_n = 0.5$]{\label{fig:kn_0.5}\includegraphics[width=0.95\linewidth]{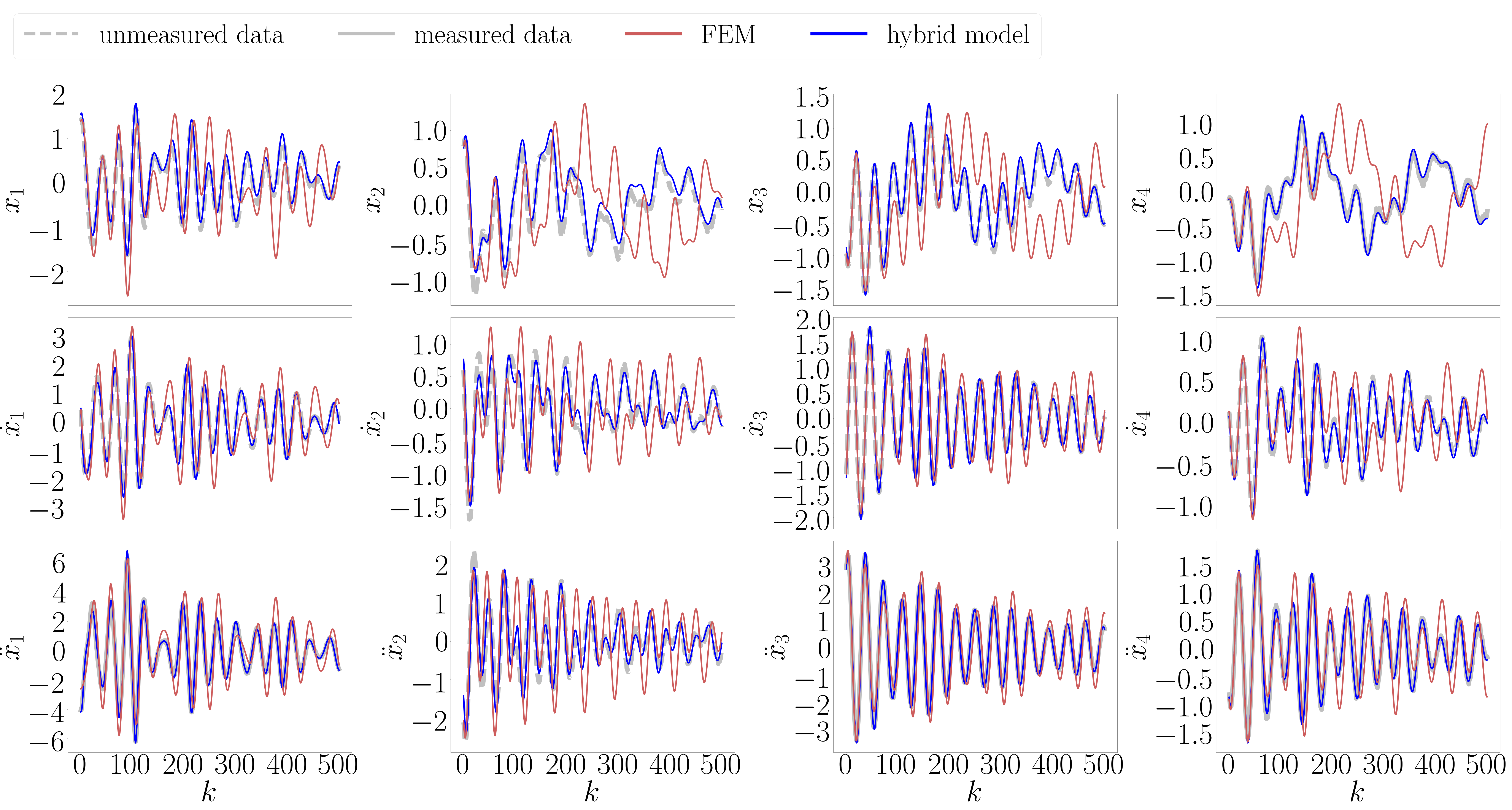}}
    \subfigure[$k_n = 1.0$]{\label{fig:kn_1}\includegraphics[width=0.95\linewidth]{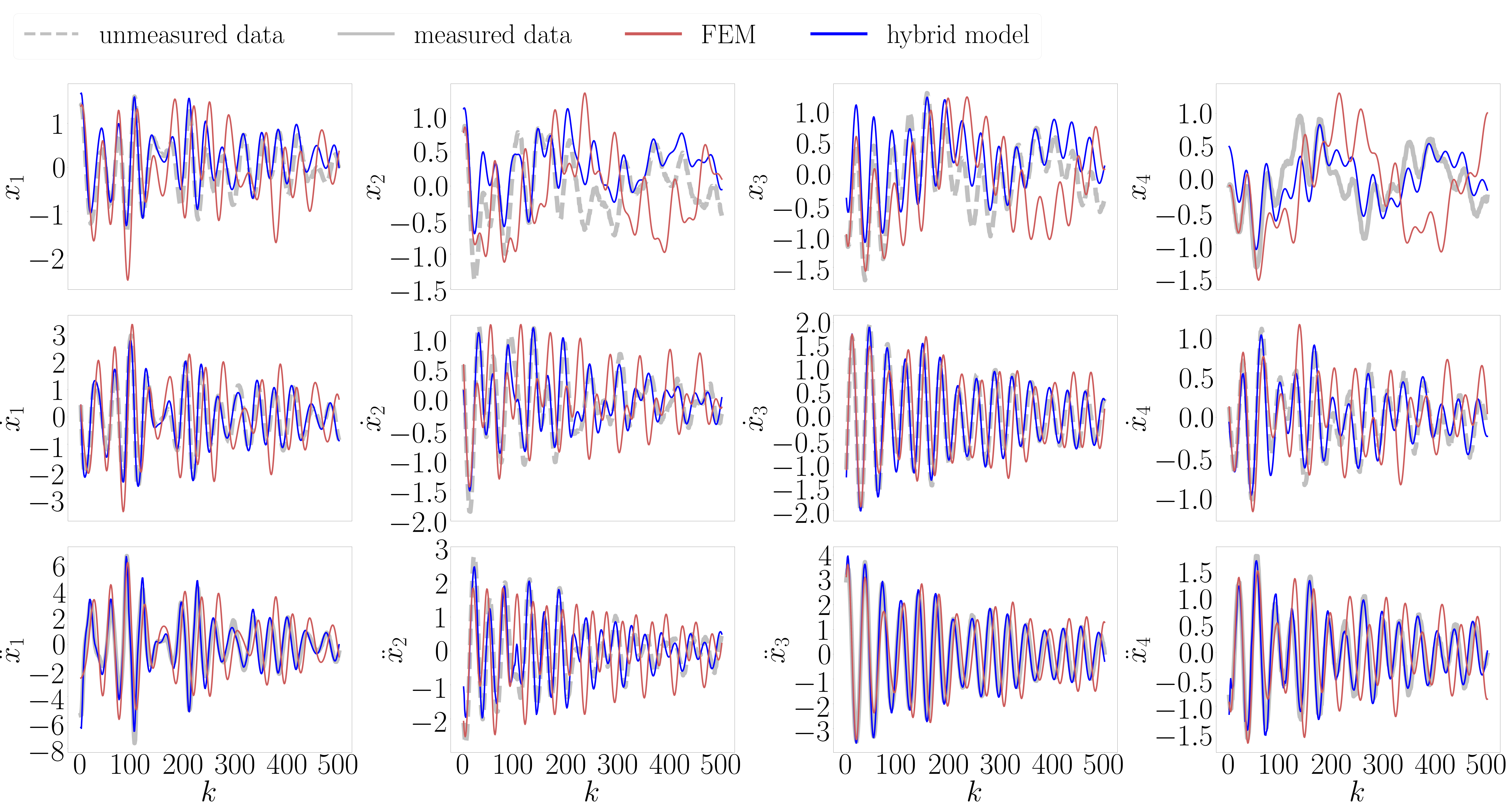}}
    \caption{Recovered full-order response for the testing data set (only $\ddot{x}_1, \ddot{x}_3, \ddot{x}_4$, and $x_4$ are measured).}
        \label{fig:test_4dof}
\end{figure}

The testing results of an exemplary realization are shown in Figures \ref{fig:test_4dof} for all three cases. In this figure, the label ``FEM" indicates a linearized model of Eq.\eqref{eq:4dof} which is intentionally contaminated with 3\% noise. The label ``Hybrid model" denotes the proposed framework -- Neural Modal ODEs.   

As shown in Figure \ref{fig:test_4dof}, the ``FEM" model approximation does not well approximate the actual response This is by design, since we purposely added noise to the model in order to simulate modeling errors. The corresponding normalized root mean squared error (NRMSE) and $R^2$ for linear regression between true and predicted responses, both averaged by the dimension $12$, are shown in Table \ref{tab:metrics}. It is observed that although the model is recommended for use with linear or mildly nonlinear systems (e.g., $k_n=0.0, 0.5$), it also performs satisfactorily for the system with relatively stronger nonlinearity ($k_n = 1.0$, which is comparable to the linear stiffness $k_1 = 1.0$). This is due to the adaption ability of the learning-based term, which is supposed to compensate the inaccuracy of the latent dynamics model $f_{\text{phy}}$, as well as to account for the imperfection of the decoder $\Phi_p$. It is also understandable that when the system becomes nonlinear, the assumption that the decoder is invariant does not hold while the responses would become energy-dependant.

\begin{table}[H]
\small
\centering
\caption{Performance metrics for the numerical study}
\begin{tabular}{c ccccc}
\hline\hline
&\multicolumn{2}{c}{Neural Modal ODE} &\vline &\multicolumn{2}{c}{FEM} \\ [0.5ex]
& NRMSE & $R^2$ &\vline & NRMSE & $R^2$ \\ [0.5ex]
\hline
$k_n=0.0$ & 0.0342 & 0.9760 &\vline & 0.1549 & 0.5635 \\ [0.5ex]
$k_n=0.5$ & 0.0496 & 0.9407 &\vline & 0.1823 & 0.4240 \\ [0.5ex]
$k_n=1.0$ & 0.0584 & 0.9431 &\vline & 0.2399 & 0.1244 \\ [0.5ex]
\hline
\end{tabular}
\label{tab:metrics}
\end{table}

For $k_n = 1.0$ (Figure \ref{fig:kn_1}), the recovery performance is not as good as the other two cases. The recovered response for $x_2$, in particular, is not perfectly aligned with the measured data. 
This implies that the decoder derived from the linear portion is not close to the actual one.  Given the limited number of measurements (observations), the model returns a discrepancy with respect to the true model. 

\section{Illustration on a Model Cable-stayed Bridge}

In this section, the proposed framework is validated on a laboratory-based monitoring dataset derived from a scaled cable-stayed bridge, which was built and tested by the Research Division on Structural Control and Health Monitoring at Tongji University, China.

\begin{figure}[H] 
  \centering 
  \begin{tabular}{c}
  \includegraphics[width=14.0cm]{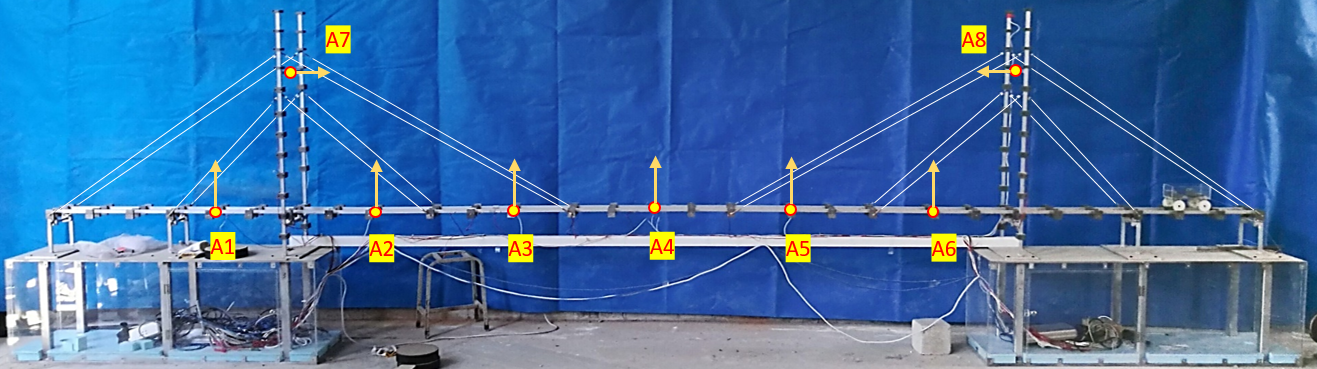} \\ (a) \\
  \includegraphics[width=14.0cm]{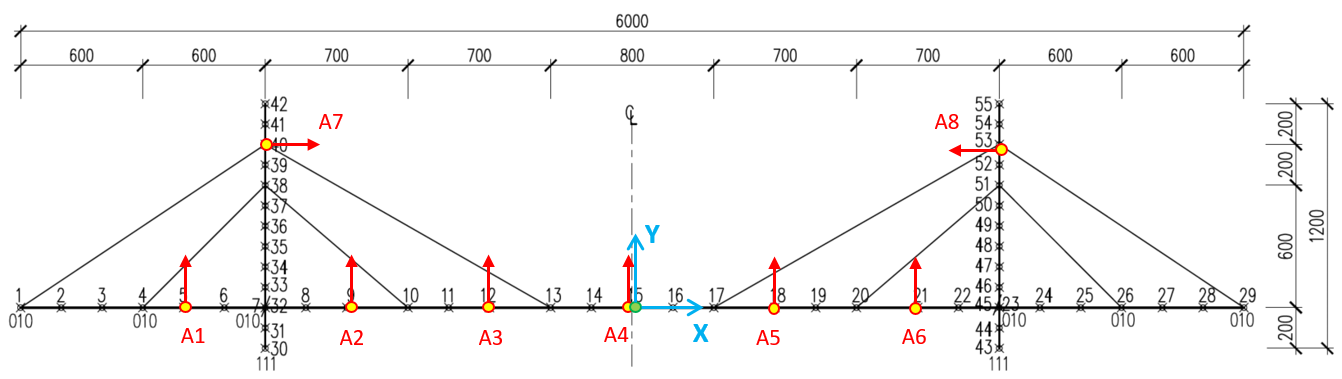} \\ (b) \\
  \end{tabular}
  \caption{Scale model cable-stayed bridge: (a) in-situ photo; (b) diagram of the finite element model (unit: mm). The eight deployed accelerometers are labelled as A1, A2,..., A8, with arrows indicating the sensing directions; the coordinate system is defined by the direction of ``X-Y" in the diagram.}
  \label{fig:bridge_model}
\end{figure}
\subsection{Experimental Setup and Data Description} \label{sec:bridge_example}
As shown in Figure \ref{fig:bridge_model}(a), this model bridge consists of one 6-meter continuous beam, two towers, and sixteen cables. The beam and towers are made of aluminum alloy, and additional metal weights are attached onto the beam and towers, ensuring that the scaled model's dynamic properties closely approximate those of the real cable-stayed bridge. 

Cable-stayed bridges are known for exhibiting geometric non-linearities. Generally speaking, the non-linear effects in cable-stayed bridges include: i) cable sag effects: cables sag because of their self-weight, resulting in variation of their axial stiffness; ii) P-delta effects: the horizontal components of cable forces bend the vertically compressed bridge pylon, introducing additional bending moments. The main girder of a cable-stayed bridge also suffers from the P-delta effects, where the bending girder is compressed by the horizontal components of cable forces; iii) large displacement effects: the displacement of the girder can be large, as the main girder of a cable-stayed bridge is mainly supported by flexible cables; thus, the small deformation assumption and linear beam theory do not apply in this scenario.

In this study, the model cable-stayed bridge exhibits nonlinearity in terms of the P-delta and large displacement effects, but these two effects are mild owing to the relatively small dimension of the scaled model. Further, the cable sag effects are negligible, as the steel cables are light. As a result, this scaled cable-stayed bridge model manifests mild nonlinearity and can be well approximated by the proposed scheme.

To measure the dynamic response of the bridge model, as highlighted in Figure \ref{fig:bridge_model}, eight MEMS (Micro-Electro-Mechanical System) accelerometers -- labeled as A1 to A8 -- are deployed on the structure, and a wired connection is used to collect the acceleration data to a digital data acquisition system. Acceleration measurements are collected at a sampling rate of 100 Hz, while the collected raw data is low-pass filtered at 30 Hz, as the dominant power in the spectrum of the raw signal lies below 30 Hz.

A ``pull-and-release" action was used to excite the bridge model. A 1 kg iron weight was hung on node 19 with a wire. When the bridge model and the weight were both stationary, the wire was abruptly cut, inducing a damped free vibration of the bridge model, and those bridge responses were recorded with the accelerometers. It is worth mentioning that the weight was hanged at the exact lateral center of the beam, so the out-of-plane vibration such as torsion was supposed to be negligible.

Five repeated tests were performed. Four of these tests were used for training, with the remaining test serving as a testing dataset.

\subsection{Finite Element Modeling}
A two-dimensional (2-D) finite element model (FEM) of the scaled bridge has been developed, in a MATLAB environment \cite{MATLAB:2019b}, which serves as the physics-based model to be adopted within the proposed deep learning framework. We consider a 2-D model as the expected motion and the deployed sensors lie within a plane. The dimension, boundary conditions, node number, coordinate system, and sensor position of the FEM are displayed in Figure \ref{fig:bridge_model}(b). Each node corresponds to three degrees-of-freedom: horizontal ($x$), vertical ($y$), and rotation. The notation ``010" in Figure \ref{fig:bridge_model}(b) signifies that vertical movement is restricted, while the horizontal and rotational movement are free (nodes 1, 4, 7, 23, 26, 29 are of this case); ``111" signifies that all the three possible DOFs are restricted (nodes 30 and 43 are of this case). The beam and towers are simulated using the Euler–Bernoulli beam element, and the cables are modeled with the tension-only truss element. The total number of degrees-of-freedom of the FEM model is 153, after applying the boundary conditions.

\begin{figure}[H]
    \centering
    \includegraphics[width= .99\linewidth]{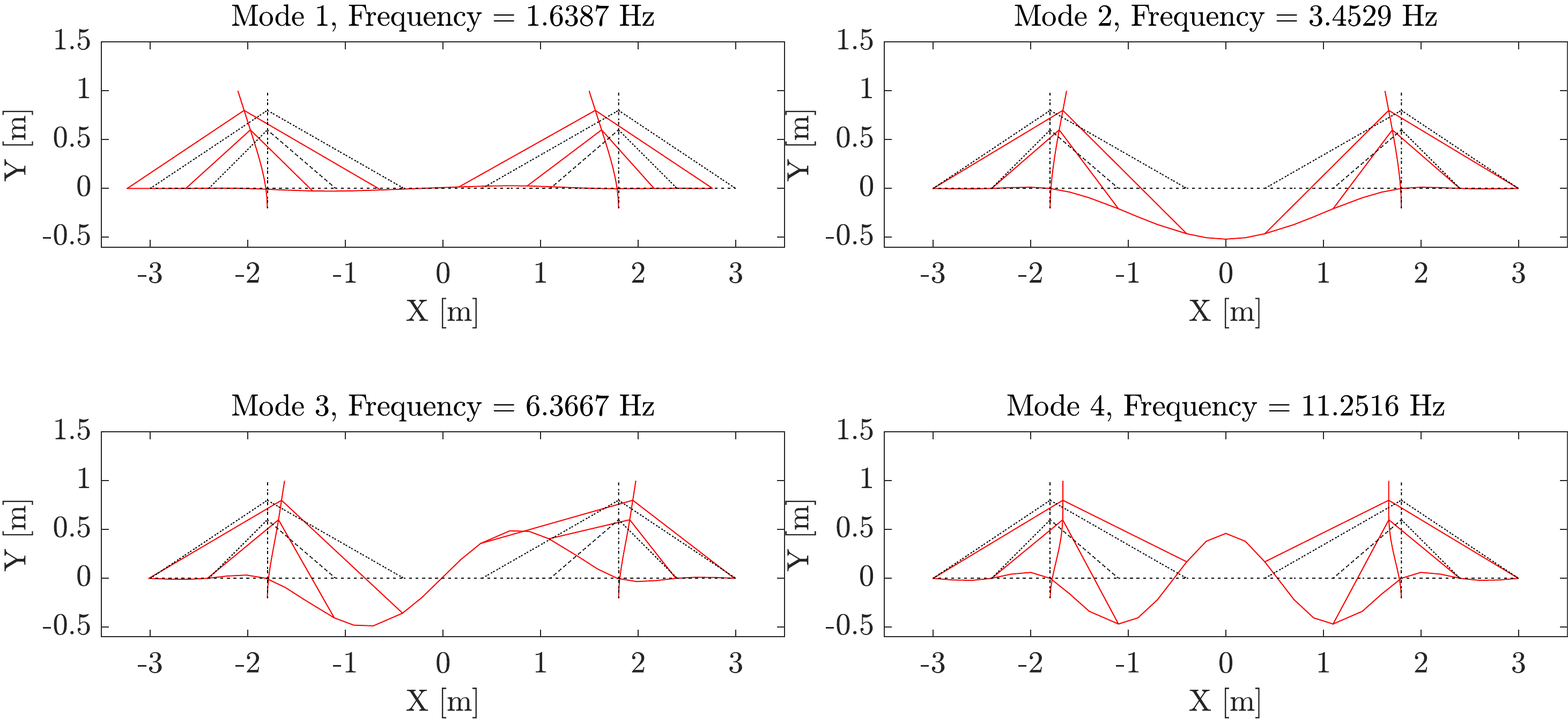}
    \caption{The first four mode shapes (denoted by red lines) derived from the eigenvalue analysis of the FEM model.}
    \label{fig:mode_shapes}
\end{figure}

Eigen-analysis is performed on the FEM model of this bridge, and the first four mode shapes ($\phi_1$ to $\phi_4$) and corresponding frequencies ($\frac{\omega_1}{2\pi}$ to $\frac{\omega_4}{2\pi}$) are shown in Figure \ref{fig:mode_shapes}, which are a horizontal drifting mode (1.6387 Hz), followed by three vertical bending modes (3.4529 Hz, 6.3667 Hz, and 11.2516 Hz).

\subsection{Model Implementation}

In this example, for modeling the latent dynamics via Eq.\eqref{eq:Pi-Neural}, we adopt the first 10 modes to construct the latent dynamics, i.e., $p=10$; $\bm{\Lambda} = \text{diag}(\omega_1^2, \omega_2^2, ..., \omega_{10}^2)$ and $\bm{\Gamma} = \text{diag}(2\xi_1\omega_1 , 2\xi_2\omega_2 , ..., 2\xi_{10}\omega_{10})$. The decoder $\Phi_p = [\phi_1, \phi_2, ..., \phi_{10}] \in \mathbb{R}^{153\times 10}$, mapping the lower dimensional latent variables back to the full order of 153; $\phi_1$ to $\phi_{10}$ are the first 10 mode shapes.

To train the model, the channels A1, and A3-A8 are used, while it is noted that the channel A2 is left out (considered as ``unmeasured") to be used for evaluating the performance of reconstruction, i.e., the model uses the sensor data at a few DOFs to reconstruct a full-order response. 

The data set includes multiple repeated free-vibration cases of the bridge, introduced by cutting a string that hangs a 1kg mass on node 19. The whole data set is divided into batches for training the model, and the number of time steps for each batch is equally 500. Thus, for each batch, the initial conditions are different, which is beneficial for training the encoder of the model. In addition, we normalize the measured acceleration across from A1 to A8, so that the maximum amplitude is 1.0, which is unitless. The details of the involved neural networks are listed in Table \ref{tab:NNs}.
\begin{table}[H]
 \small
    \centering
    \caption{Implementation details for the experimental study}
    \begin{tabular}{c c c c c}
   
    \hline\hline
      & \multicolumn{2}{c}{Encoder} & Modeling latent dynamics & Decoder \\
        \hline
    &  \makecell{RNN($\textbf{x}_{0:n_t}$)\\\small{$n_t = 10$}} &  MLP($\textbf{x}_0$)   &    $\text{NN}(\textbf{z})$ in Eq.\eqref{eq:Pi-Neural} & $\Phi_p$ \\
  \hline
\makecell{no. of \\hidden layers} & 1 & 2 & 2 &\multirow{2}{*}{\makecell{invariant,\\\small{$\Phi_p = [\phi_1, \phi_2, ..., \phi_{10}]$}}   }\\  
\makecell{no. of neurons in \\each hidden layer}  & 128 & 128 & 128 & \\  
  \hline
    \end{tabular}
    \label{tab:NNs}
\end{table}

\subsection{Results}
\begin{figure}[H]
    \centering
    \includegraphics[width= .82\linewidth]{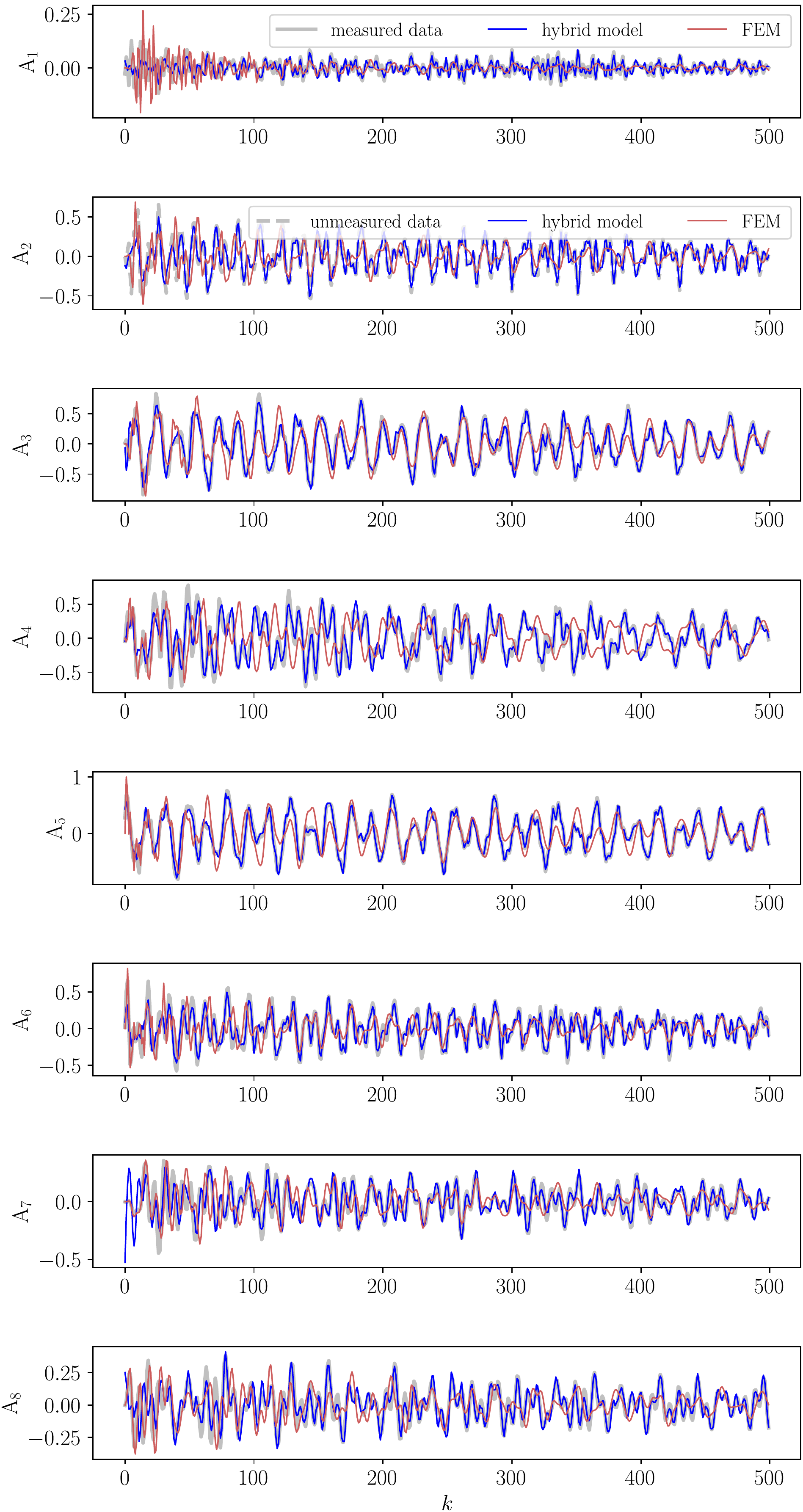}
    \caption{Comparisons of acceleration responses prediction between actual measurements, the proposed hybrid model (Neural Modal ODEs), and FEM model (A1 - A8 are normalized unitless data with maximum value of 1; the horizontal axis $k$ denotes the time step).}
    \label{fig:acc_pred}
\end{figure}

Once the model has been trained, the trained model is used for predicting the structural responses.
The corresponding predictions of acceleration A1 - A8 are shown in Figure \ref{fig:acc_pred}, denoted by the blue lines. This prediction is compared with the actual measurements in grey color and predictions by the FEM models in red color. One can see that the FEM model offers satisfactory results, while some channel predictions are out of phase and fail to accurately follow the actual measurement, most possibly due to the inaccurate modeling of damping (this can be clearly observed in the A4 channel). The prediction from the proposed hybrid model is evidently more accurate than the FEM model, almost aligning with the actual measurements.

It is noted that the data of the A2 channel is unmeasured and not used for training the hybrid model, denoted by dashed grey lines. The prediction shown in the A2 plot comes from the full-order reconstructed responses. One can see that the reconstruction of A2 still highly agrees with the actual data, even though it is not used for the training.

\begin{figure}[H]
    \centering
    \includegraphics[width= 1.0\linewidth]{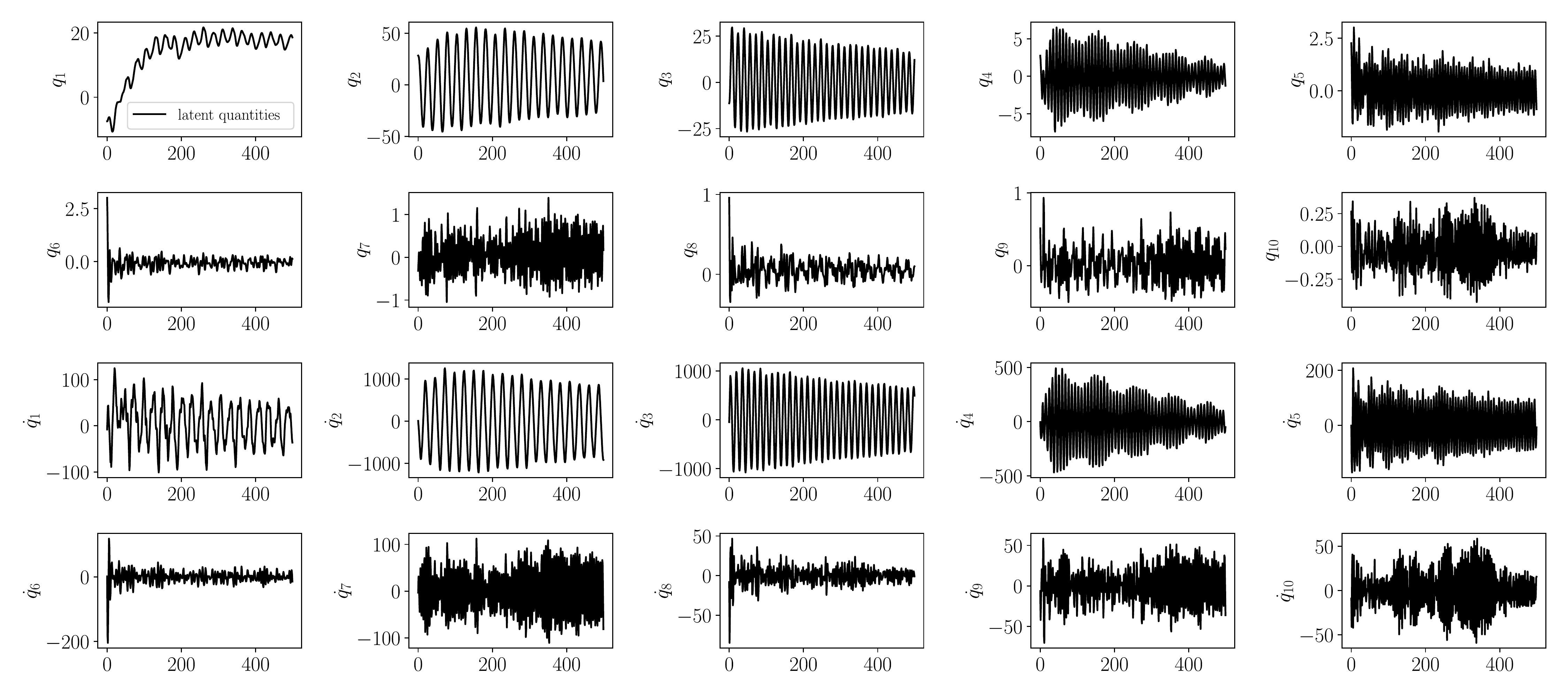}
    \caption{The learned latent variables $\textbf{q} = [q_1, q_2, ..., q_{10}]^T$ and $\dot{\textbf{q}} = [\dot{q}_1, \dot{q}_2, ..., \dot{q}_{10}]^T$. (The $x-\text{axis}$ in each subplot is time step)}
    \label{fig:latent}
\end{figure}

\begin{figure}[H]
    \centering
    \includegraphics[width= 0.75\linewidth]{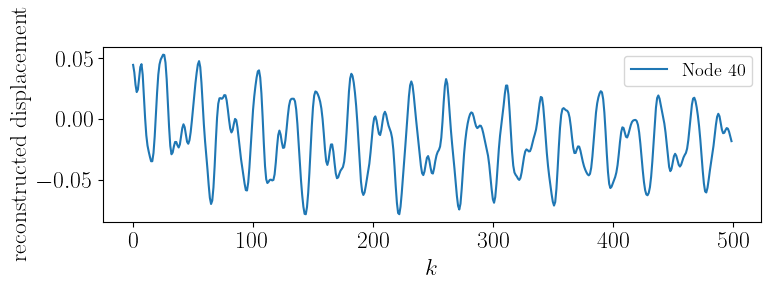}
    \caption{Reconstructed displacement at Node 40 (noted that in this example, as the acceleration is normalized, the reconstructed displacement is only a scaled version).}
    \label{fig:Node40}
\end{figure}

\begin{figure}[H]
    \centering
    \includegraphics[width= .95\linewidth]{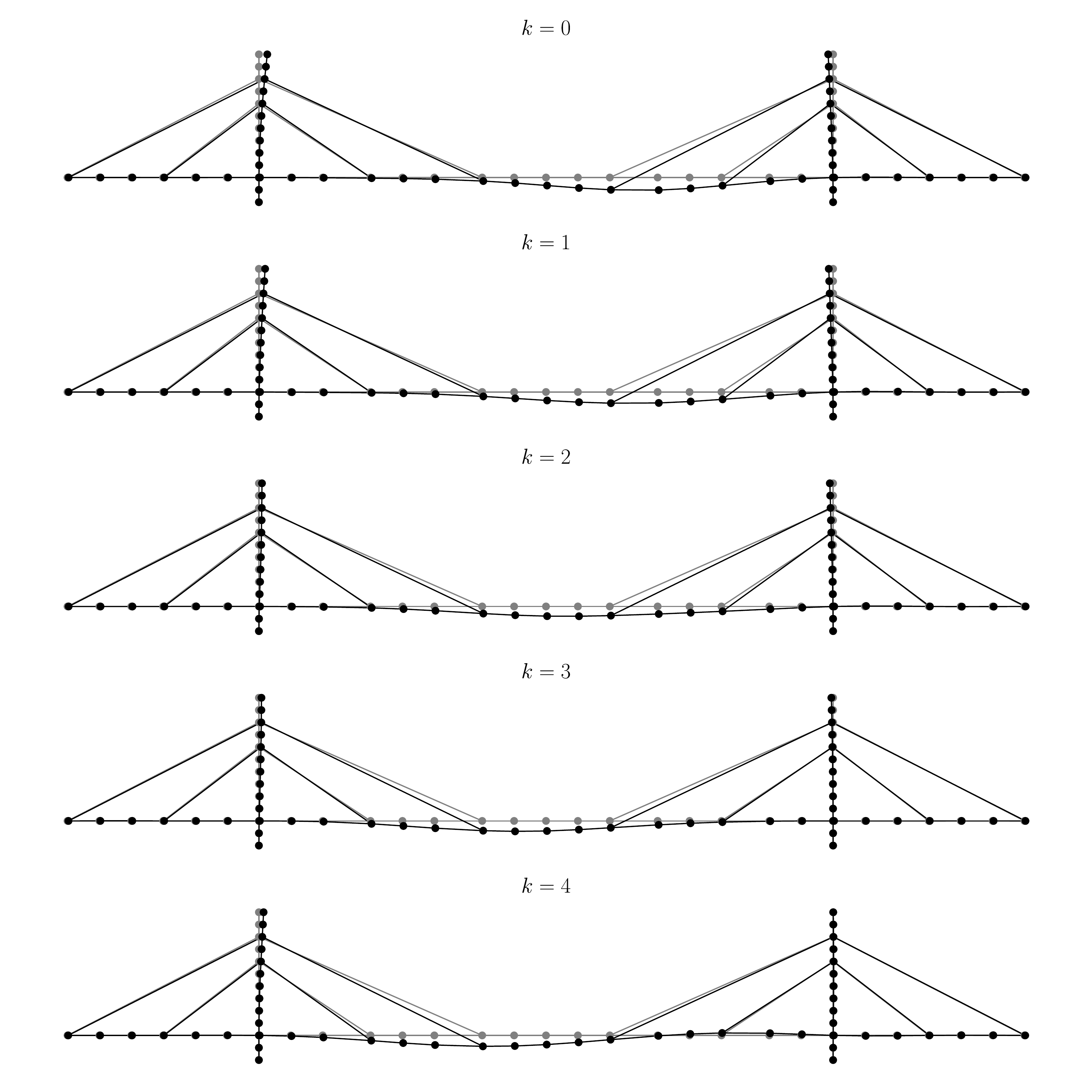}
    \caption{The reconstruction of the full-order displacement responses (using the first 5 snapshots as an example).}
    \label{fig:dis_full}
\end{figure}

\begin{figure}[H]
    \centering
    \includegraphics[width= 1\linewidth]{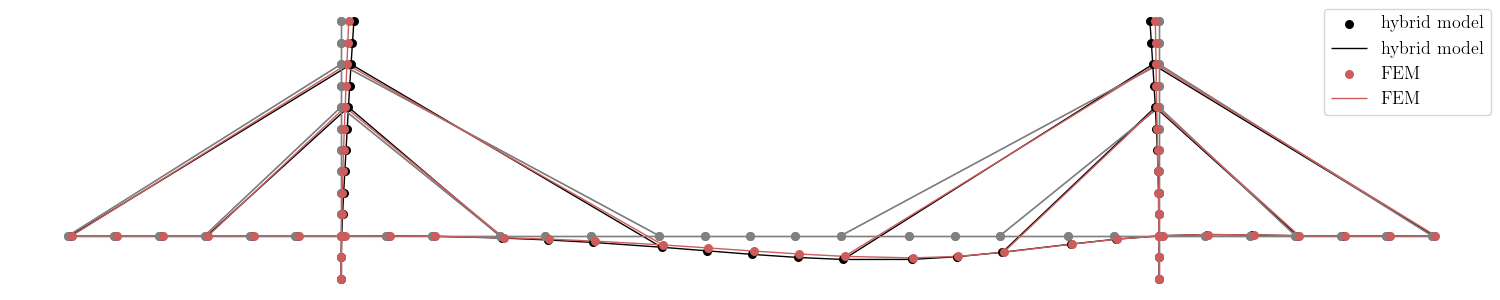}
    \caption{The comparisons of the initial deformation of the bridge between the reconstruction from the proposed hybrid model (Neural Modal ODEs) and FEM model.}
    \label{fig:hybrid_fem}
\end{figure}

Figure \ref{fig:latent} shows the corresponding learned time history of latent variables $\textbf{q} = [q_1, q_2, ..., q_{10}]^T$ and $\dot{\textbf{q}} = [\dot{q}_1, \dot{q}_2, ..., \dot{q}_{10}]^T$, related to displacement and velocity in modal coordinates, respectively. It is observed that: (i) $q_1$ to $q_{10}$ retains the order from low-frequency to high-frequency, that we impose in the physics-informed term. In addition, these ``modes" are near mono-frequent, almost preserving the decoupled structure; (ii) by examining the amplitude of the latent variables, we are able to tell the contribution level of each mode. $q_1$, $q_2$, $q_3$, and $q_4$ (hence, $\dot{q}_1$, $\dot{q}_2$, $\dot{q}_3$, and $\dot{q}_4$) have the highest amplitudes, dominating the vibration, while the amplitudes of other higher modes are much smaller (close to residuals). This is well understandable since for this free vibration, only the first several modes are fully excited while others are weakly present; (iii) it is interesting to see that $q_1$ initiates from a value and then oscillates around an equilibrium which is not close to zero. $q_1$ is the modal displacement corresponding to the first horizontal drifting mode, which can only be picked up by A7 and A8 in the horizontal direction, at Nodes 40 and 53. We show that, in Figure \ref{fig:Node40} as an example, after the decoder, the reconstructed displacement at Node 40 retains a reasonable vibration: initiating from a value and then oscillating around zero.
As stated in Eq.\eqref{eq:emission}, in this trained model, one has the flexibility of reconstructing different types of responses. For example, we reconstruct the full-order displacement responses via $\textbf{x}_t^{\text{full}} = \Phi_p(\textbf{q}_t)$. Figure \ref{fig:dis_full} shows five consecutive snapshots of the full-order reconstructed displacements away from the equilibrium position, and a more intuitive video is provided in the auxiliary files. It is observed that the reconstruction preserves the legitimate spatial relationships between each node, due to the reason that the decoder is imposed by invariant normal modes. We also did an experiment using $\Phi_p + \text{NN}$ (normal modes added by a trainable neural network to consider the imperfection of the normal modes) as a decoder. However, we find and conclude that this is not an appropriate decoder since the learned NN breaks the inherent spatial relationship between each node.
Since in this data set no reliable displacement measurements were achieved, in order to validate the accuracy of reconstructed full-order displacement from limited acceleration data, we compare the initial deformation ($k=0$) with the one derived by the FEM model. The comparison result shown in Figure \ref{fig:hybrid_fem} indicates that the reconstructed displacement from the measured acceleration data highly agrees with the computed one by the FEM model. Thus, it is valid to see that the proposed hybrid model is capable of spatially extrapolating the dynamics and also of reconstructing other types of responses from a certain type of measurement (for example, in this study case, the displacement and velocity are successfully reconstructed from the acceleration).

\section{Conclusions} 
In this paper, we propose a framework for integrating physics-based modeling with deep learning for modeling large civil/mechanical dynamical systems. The framework couples a dynamical variational autoencoder with a Physics-informed Neural ODE scheme. The autoencoder encodes a limited amount of sensed data into an estimate of the initial conditions of the latent space. This allows for the construction of a generative model which aims at predicting the latent system dynamics via a learned Physics-informed Neural ODE. The predicted dynamic response is then mapped back onto the measured physical space via an invariant decoder, which is effectuated on the basis of the eigenmodes derived from a physics-based model. The framework assimilates physics-related features from a physics-based model into a deep learning model, to yield a learned generative model, which is not eventually data-dependent and leads to an interpretable architecture. The delivered models are able to reconstruct the full field structural response, meaning response in unmeasured locations, given limited sensing locations. Future work will investigate boosting the decoder via assimilation of a Bayesian neural network.

\section{Discussions}
We want to further clarify that the extrapolation capability cannot be guaranteed if the dynamic regime differs significantly from the training data we used to train the model, which is also typically the limitation of most deep learning methods. From the numerical study, it can be observed that the proposed framework is able to capture unseen scenarios, when these do not excite a significantly higher level of non-linearity. This is why, we comment on the framework being applicable for mildly nonlinear systems, implying that in presence of sever nonlinearity the extrapolation potential is limited.


\section*{Data availability statement}

A demonstrative code (Section \ref{sec:num_exp}) that implements the proposed method is openly available at \href{https://github.com/zlaidyn/Neural-Modal-ODE-Demo}{https://github.com/zlaidyn/Neural-Modal-ODE-Demo}.

\section*{Funding statement}
The research was conducted at the Singapore-ETH Centre, which was established collaboratively between ETH Zurich and the National Research Foundation Singapore. This research is supported by the National Research Foundation, Prime Minister’s Office, Singapore under its Campus for Research Excellence and Technological Enterprise (CREATE) programme.

\section*{Competing Interests}

The authors declared no potential conflicts of interest with respect to the research, authorship, and publication of this article.

\section*{Author Contributions}
\begin{itemize}
    \item Zhilu Lai: Conceptualization (Lead),
Data curation (Supporting),
Formal analysis (Lead),
Investigation (Lead),
Methodology (Lead),
Software (Lead),
Validation (Lead),
Visualization (Lead),
Writing – original draft (Lead),
Writing – review and editing (Lead)
\smallskip
\item Wei Liu: Formal analysis (Supporting)
Investigation (Supporting),
Software (Supporting),
Validation (Supporting),
Writing – original draft (Supporting)
\smallskip
\item  Xudong Jian:
Data curation (Supporting)
\smallskip
 \item Kiran Bacsa:
Software (Supporting)
\smallskip
\item  Limin Sun:
Data curation (Lead)
\smallskip
\item  Eleni Chatzi:
Conceptualization (Supporting),
Investigation (Supporting),
Methodology (Supporting),
Project administration (Lead),
Resources (Lead),
Supervision (Lead),
Validation (Equal),
Writing – original draft (Supporting),
Writing – review and editing (Equal)
\end{itemize}

\newpage
\printbibliography


\newpage
\appendix
\section{Gradients for Linear Approximation Cases}\label{appendix_A}
As an illustration, we consider the case where a linear approximation is adopted as the physics-based model, i.e., $f_{\bm{\theta}}(\textbf{z})=\textbf{A}\textbf{z}+f_\text{NN}(\textbf{z})$, and the decoder $\Phi_p$ is simply an identity matrix of appropriate dimension. Following the derivation of gradients with repect to $\bm{\theta}$ and $\textbf{z}(t_0)$ given by \cite{chen2018neural}, the gradients under the physics-informed regime can be expressed as the solution of the following differential equation:
\begin{equation}\label{eq:gradients}
\begin{split}
    \frac{d\textbf{a}_\text{aug}(t)}{dt}=
    -\begin{bmatrix}
    \frac{\partial f}{\partial \textbf{z}}\\
    \frac{\partial f}{\partial \bm{\theta}}
    \end{bmatrix} \textbf{a}(t)
    &=-\begin{bmatrix}
    \textbf{A}\\
    \textbf{0}
    \end{bmatrix} \textbf{a}(t)
    -\begin{bmatrix}
    \frac{\partial f_\text{NN}}{\partial \textbf{z}}\\
    \frac{\partial f_\text{NN}}{\partial \bm{\theta}}
    \end{bmatrix} \textbf{a}(t)\\
    &=-\textbf{A}_\text{aug}\textbf{a}_\text{aug}(t)-\begin{bmatrix}
    \frac{\partial f_\text{NN}}{\partial \textbf{z}} & \textbf{0}\\
    \frac{\partial f_\text{NN}}{\partial \bm{\theta}} & \textbf{0}
    \end{bmatrix} \textbf{a}_\text{aug}(t),
\end{split}
\end{equation}
where $\textbf{A}_\text{aug}=\begin{bmatrix} \textbf{A} & \textbf{0} \\ \textbf{0} & \textbf{0} \end{bmatrix}$, $\textbf{a}_\text{aug}= \begin{bmatrix} \textbf{a}\\ \textbf{a}_{\bm{\theta}} \end{bmatrix}$, $\textbf{a}(t)=\frac{dL}{d\textbf{z}(t)}$, $\textbf{a}_{\bm{\theta}}(t)=\frac{dL}{d\bm{\theta}(t)}$.
Then the gradients can be approximately given by
\begin{equation}
    \textbf{a}_\text{aug}(t_0)=\textbf{a}_\text{aug}(t_1)\text{exp}({\textbf{A}_\text{aug}}\Delta_t)+F_\text{NN}(t_1),
\end{equation}
where the first term is an approximate solution obtained from linear physics-based portion of \ref{eq:gradients} and the second term $F_\text{NN}$ accounts for the difference between the linear approximation and the true solution. Suppose the training time steps are $t=t_0,t_1,...,t_N$, then we can repeat the process and the gradient obtained through back-propagation is:
\begin{equation}
    \textbf{a}_\text{aug}(t_0)=\textbf{a}_\text{aug}(t_N)\text{exp}({\textbf{A}_\text{aug}}N\Delta_t)+\sum_{i=1}^{N}F_\text{NN}(t_i).
\end{equation}
The first term of the R.H.S. is brought by the linearized physics-based model and it can be directly back-propagated, while only the discrepancy terms $\sum_{i=1}^{N}F_\text{NN}(t_i)$ need to be estimated, which makes the estimated gradients also an approximation to the real ones. As a result from this, the combined gradients are restricted in a regime that is closer to the true function's gradients.

\end{document}